%% file: main.tex
\documentclass[nohyperref]{article}

\usepackage[table]{xcolor}
\usepackage{microtype}
\usepackage{graphicx}
\usepackage{subfigure}
\usepackage{tikz}
\usepackage{booktabs} 
\usepackage{enumitem}

\definecolor{mydarkblue}{RGB}{0, 0, 139}

\usepackage[backref=page, colorlinks=true,
    linkcolor=mydarkblue,
    citecolor=mydarkblue,
    filecolor=mydarkblue,
    urlcolor=mydarkblue]{hyperref}

\usepackage{multirow, graphbox}
\usepackage{tabularx}
\usepackage{pifont}
\usepackage{soul}

\include{macros}
\graphicspath{ {figs/} }

\newcommand{\ghnurl}{\url{https://github.com/SamsungSAILMontreal/ghn3}}

\usepackage[accepted]{icml2023}

\usepackage{amsmath}
\usepackage{amssymb}
\usepackage{mathtools}
\usepackage{amsthm}
\usepackage{xspace}

\usepackage[capitalize,noabbrev]{cleveref}

\usepackage[disable,textsize=tiny]{todonotes}

\icmltitlerunning{
Can We Scale Transformers to Predict Parameters of Diverse ImageNet Models?}

\begin{document}

\twocolumn[
\icmltitle{Can We Scale Transformers to Predict Parameters of Diverse ImageNet Models?}
\begin{icmlauthorlist}
\icmlauthor{Boris Knyazev}{sail}
\icmlauthor{Doha Hwang}{sait}
\icmlauthor{Simon Lacoste-Julien}{sail,udem,cifar}
\end{icmlauthorlist}

\icmlaffiliation{sail}{Samsung - SAIT AI Lab, Montreal}
\icmlaffiliation{udem}{Mila, Université de Montreal}
\icmlaffiliation{sait}{Samsung Advanced Institute of Technology (SAIT), South Korea}
\icmlaffiliation{cifar}{Canada CIFAR AI Chair}
\icmlcorrespondingauthor{Boris Knyazev}{borknyaz@gmail.com}

\icmlkeywords{Machine Learning, ICML, HyperNetworks, Transformers, ImageNet, Pretraining, Graphs}

\vskip 0.3in

\begin{center}
 \vspace{-12pt}
 \textcolor{violet}{\ghnurl}
 \vspace{0.12in}
\end{center}

]

\printAffiliationsAndNotice{}  

\begin{abstract}
Pretraining a neural network on a large dataset is becoming a cornerstone in machine learning that is within the reach of only a few communities with large-resources. We aim at an ambitious goal of democratizing pretraining. Towards that goal, we train and release a single neural network that can predict high quality ImageNet parameters of other neural networks. By using predicted parameters for initialization we are able to boost training of diverse ImageNet models available in PyTorch. When transferred to other datasets, models initialized with predicted parameters also converge faster and reach competitive final performance.
\end{abstract}

\section{Introduction}
\label{sec:intro}

Training a neural network $f$ initialized with parameters $\w$ is typically done by running a stochastic gradient descent (SGD) optimization algorithm on a dataset $\domain$:
\setlength{\belowdisplayskip}{6pt}
\setlength{\abovedisplayskip}{6pt}
\begin{equation}
    \label{eq:sgd}
    {\w}^* = \text{SGD}(f, \w, \domain).
\end{equation}
Novel neural architectures, e.g. Vision Transformer~\cite{dosovitskiy2020image}, are usually first pretrained by Eq.~\eqref{eq:sgd} on some large $\domain$ such as ImageNet~\cite{russakovsky2015imagenet} or, some in-house data such as JFT-300M, and then transferred to other downstream tasks~\cite{kornblith2019better,zhai2019large,dosovitskiy2020image}. With the growing size of networks $f$ and with multiple runs needed (e.g. for hyperparameter tuning), the cost of pretraining is becoming unsustainable~\cite{strubell2019energy,thompson2020computational,zhai2022scaling}. Therefore, pretraining is becoming one of the key factors increasing the gap between a few privileged communities (often big industries) and many low-resource communities (often academics and small companies).

\begin{figure}[ht]
\vskip 0.2in
\begin{center}
\centerline{\includegraphics[width=\columnwidth]{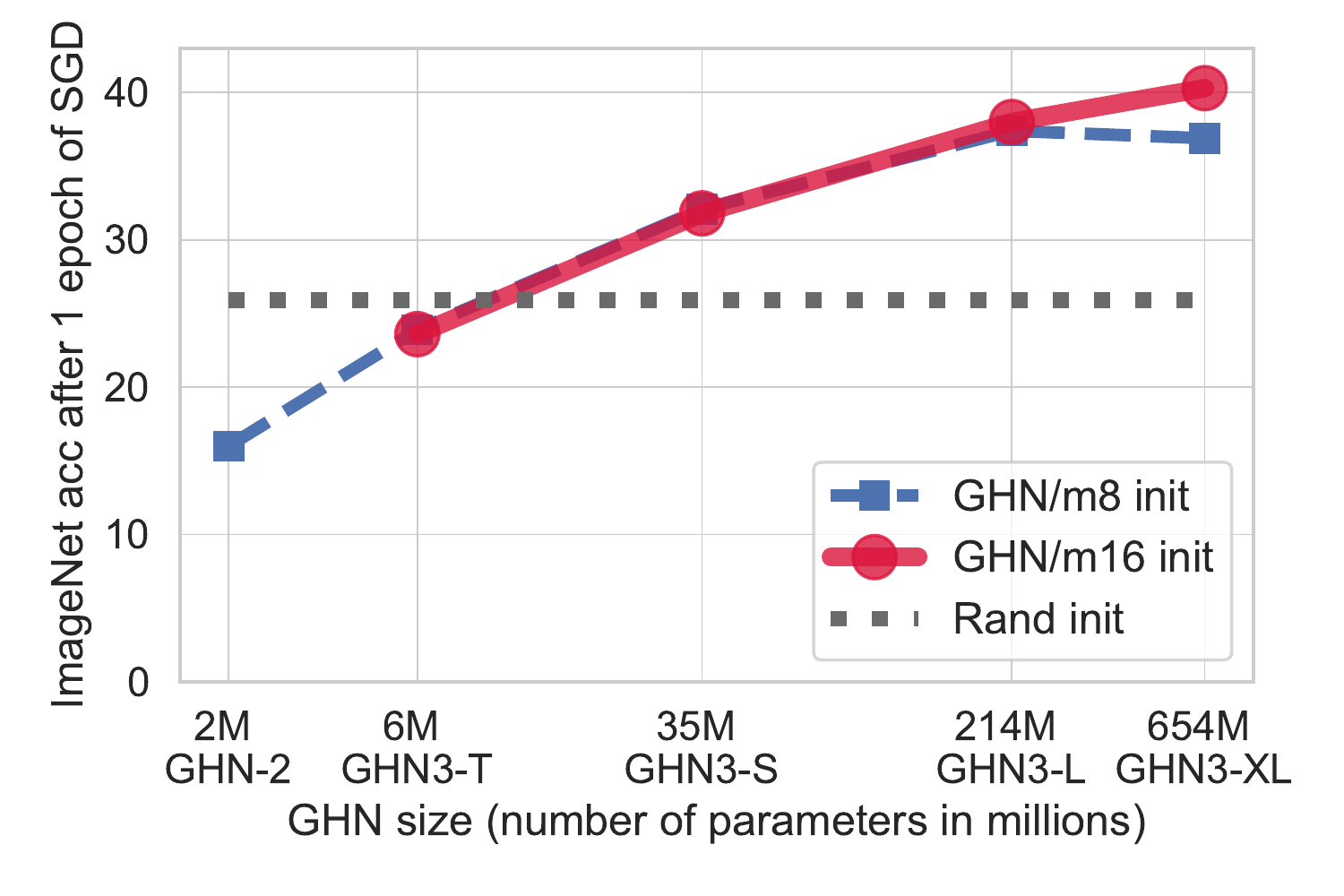}}
\vspace{-10pt}
\caption{We introduce \ghnthree models of a significantly larger scale and larger training meta-batch size ($m$) compared to \ghntwo~\cite{knyazev2021parameter}. Scaling up GHNs leads to consistent improvements in the quality of predicted parameters when used as initialization on ImageNet. This plot is based on the accuracies of the \textsc{PyTorch}-10 models in Table~\ref{tab:bench_imagenet}.}
\label{fig:scaleup-teaser}
\end{center}
\vskip -0.3in
\end{figure}

\begin{figure*}[ht]
\vskip 0.02in
\begin{center}
\newcommand\width{0.17\textwidth}
\setlength{\tabcolsep}{10pt}
\begin{tabular}{lccc}
     
     \small \textsc{\textbf{Initialization:}} & \textbf{\ghntwo} & \textbf{\ghnxl/m16 (ours)} & \textsc{\textbf{Rand init}}\\
     \midrule
     
     & \includegraphics[width=\width, align=c, trim={6.35cm 2.0cm 35.8cm 2.1cm}, clip]{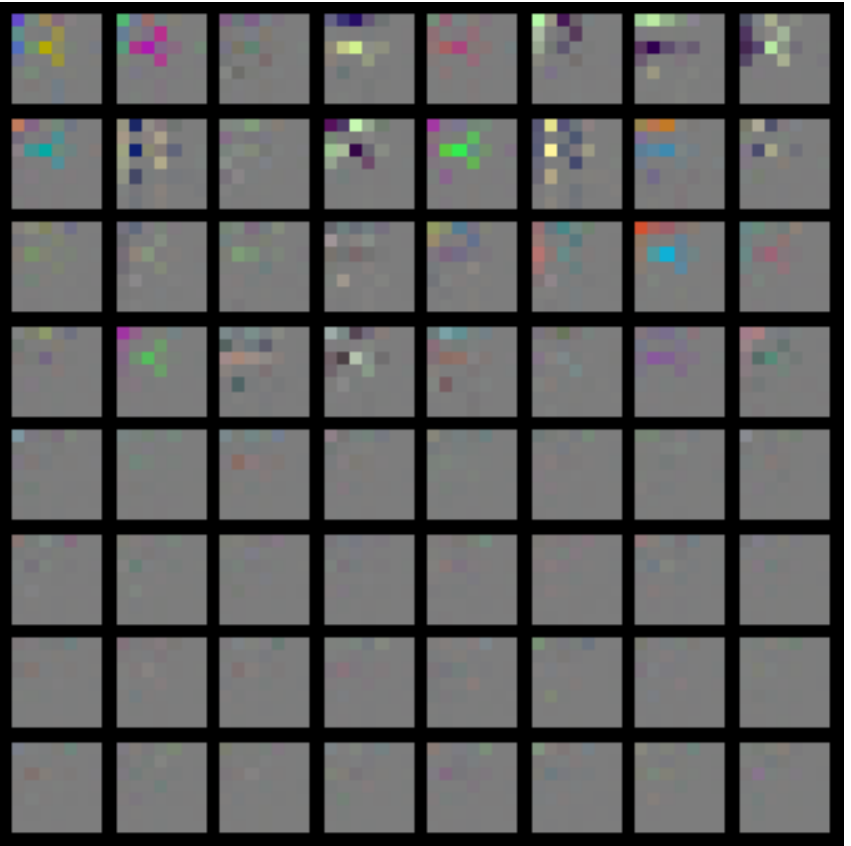} & 
     
    \includegraphics[width=\width, align=c, trim={6.35cm 2.0cm 35.8cm 2.1cm}, clip]{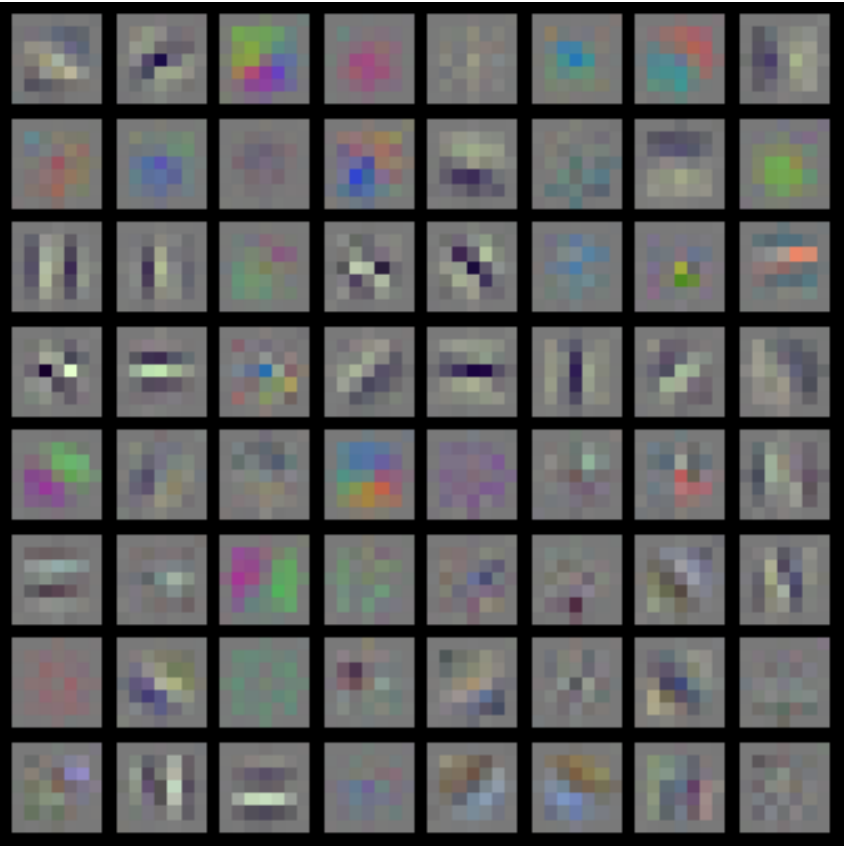} &
     
     \includegraphics[width=\width, align=c, trim={6.35cm 2.0cm 35.8cm 2.1cm}, clip]{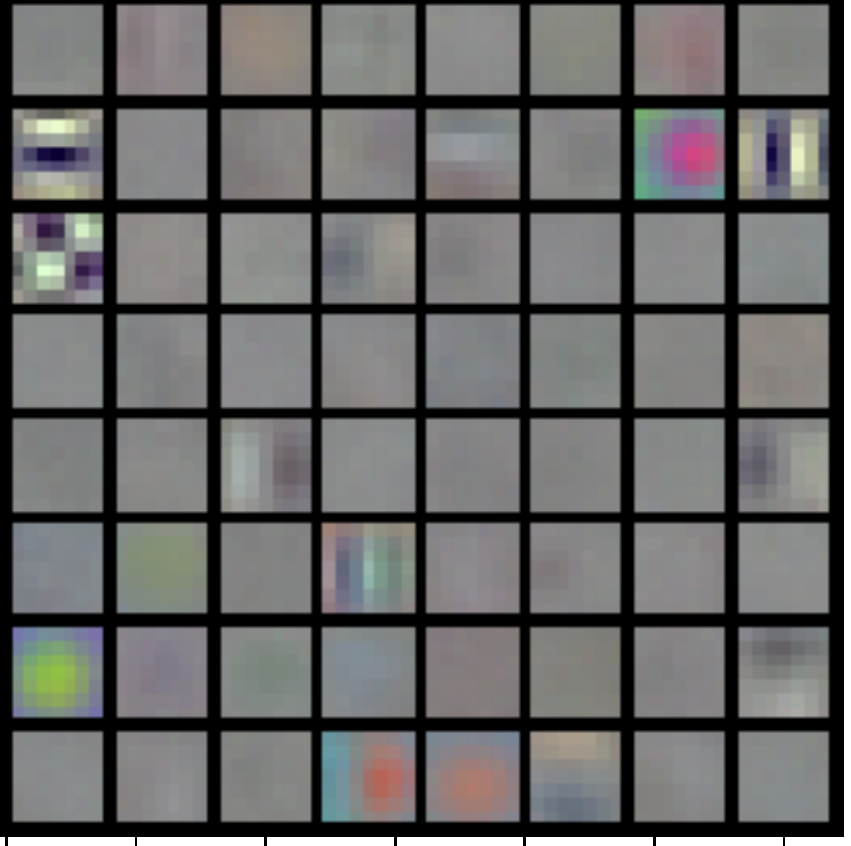} \Tstrut \\
     
     \small Parameters trained for: & no training &  no training & 1 epoch \\
     
     \small Time (sec): & 0.9 on \textbf{CPU} & 1.1 on \textbf{CPU} & $4\times 10^3$ on \textbf{GPU} \\
     
     \small ImageNet acc: & 1.1\% & \textbf{20.0\%} & 18.2\% \\
     \\
     
     & \includegraphics[width=\width, align=c, trim={6.35cm 2.1cm 35.8cm 2.0cm}, clip]{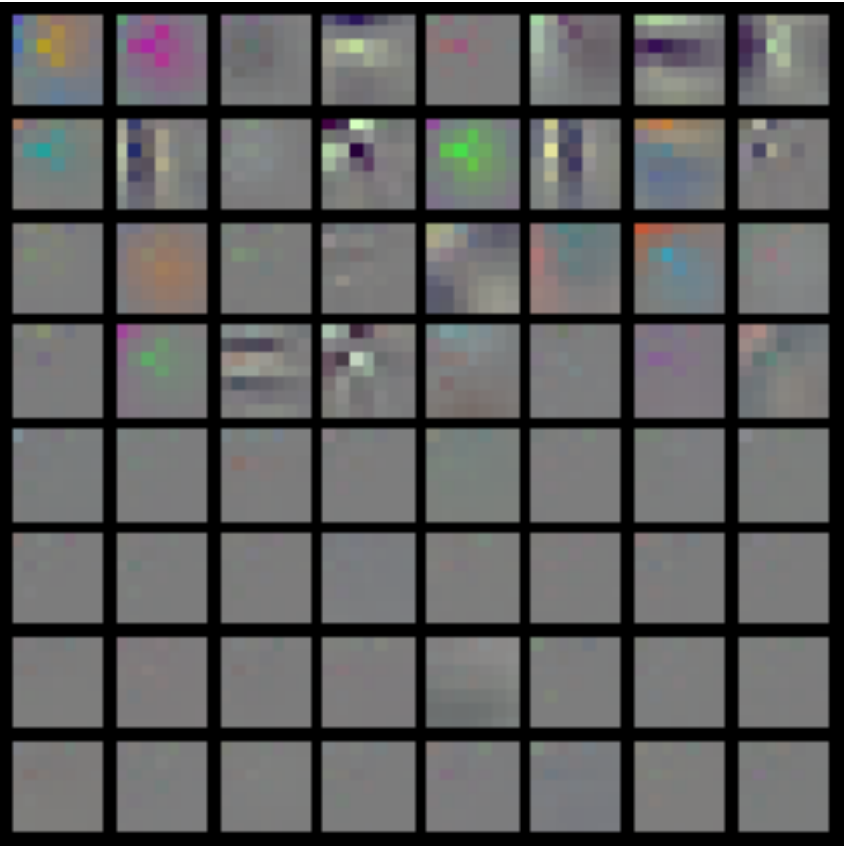} &
     
     \includegraphics[width=\width, align=c, trim={6.35cm 2.1cm 35.8cm 2.0cm}, clip]{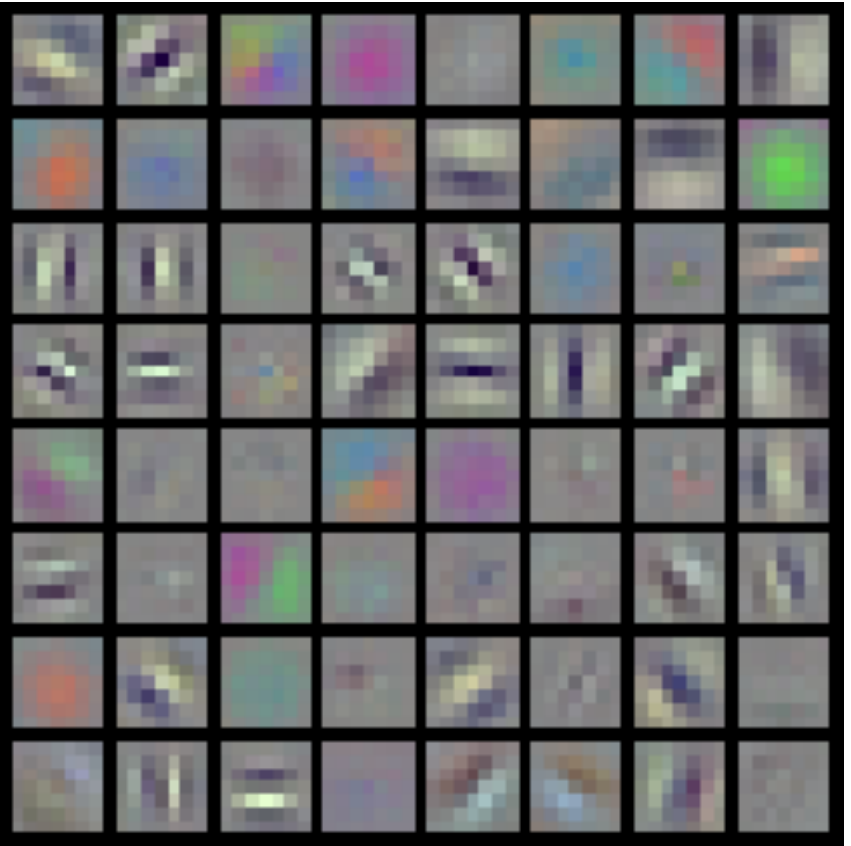} &
     
     \includegraphics[width=\width, align=c, trim={6.35cm 2.1cm 35.8cm 2.0cm}, clip]{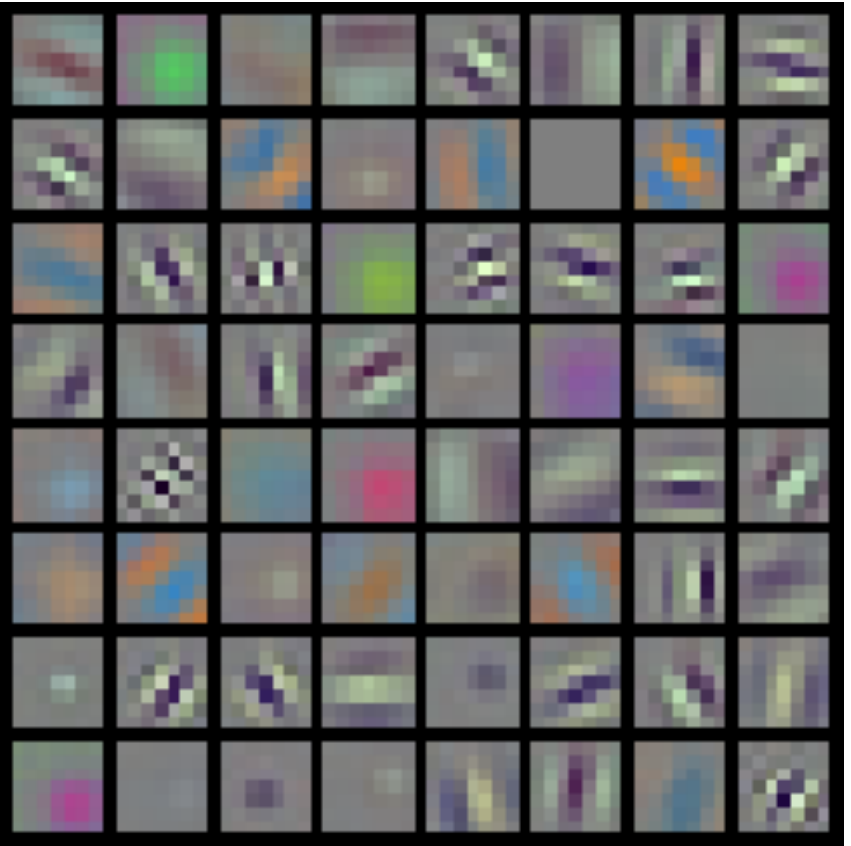} 
     \\
     
     \small Parameters trained for: &  1 epoch & 1 epoch & 90 epochs \\
     
     \small Time (sec): & $4\times 10^3$ on \textbf{GPU} & $4\times 10^3$ on \textbf{GPU} & $6\times 10^5$ on \textbf{GPU} \\
     
     \small ImageNet acc: & 14.6\% & \textbf{43.7\%} & 76.1\% \\
     
\end{tabular}
\vspace{-3pt}
\caption{ResNet-50 first layer parameters predicted with the baseline \ghntwo, our \ghnxl/m16 and optimized with SGD on ImageNet for 1 and 90 epochs.
For the GHNs, ResNet-50 is an unseen architecture, i.e. not present in the training dataset of architectures~\cite{knyazev2021parameter}. 
The ``Time'' row shows a rough time estimation to obtain all the parameters of ResNet-50 (for SGD the time is based on~\citet{knyazev2021parameter}).
The ``ImageNet acc'' row shows top-1 validation accuracy on ImageNet ILSVRC-2012. Our \ghnthree predicts parameters as fast as \ghntwo, but significantly improves the quality of predicted parameters making their performance higher, fine-tuning easier and visual appearance similar to those trained with SGD for 90 epochs. See Section~\ref{sec:exper} for experiment details.}
\label{fig:resnet50-conv1}
\end{center}
\vskip -0.2in
\end{figure*}

We aim at an ambitious goal of democratizing the cost of pretraining. To do so, we follow recent works where one network (\textit{HyperNetwork}) parameterized by $\theta$ is trained to predict good parameters $\w_\text{pred}$ for unseen network architectures $f$~\cite{zhang2018graph,knyazev2021parameter,shang2022one} or datasets $\domain$~\cite{zhmoginov2022hypertransformer}.
We focus in this paper on generalization to new architectures $f$ in a large-scale ImageNet setting so that the HyperNetwork~$H_{\domain}$ is used as:\looseness-1
\begin{equation}
    \label{eq:predict}
    \w_\text{pred} = H_{\domain}(f, \theta).
\end{equation}
We use the parameters $\w_\text{pred}$ predicted on ImageNet ($\domain$) as initialization, so by fine-tuning them on $\domain$ we can reduce pretraining cost or we can transfer them to another dataset $\domain^\text{transfer}$ by fine-tuning them on $\domain^\text{transfer}$ with Eq.~\eqref{eq:sgd}:
\begin{equation}
    \label{eq:finetune}
    {\w}^* = \text{SGD}(f, \w_\text{pred}, \domain \ \text{or} \ \domain^\text{transfer}).
\end{equation}
While training $H_{\domain}$ is more expensive than training $f$ from scratch, we train $H_{\domain}$ only once and publicly release it to move towards democratized pretraining. This way, new architectures (including very large ones) designed by different research communities may be boosted ``for free'' by initializing them with $\w_\text{pred}$ by a simple forward pass through $H_{\domain}$.
Our proposed $H_{\domain}$ is built on Graph HyperNetworks (GHNs)~\cite{zhang2018graph}, in particular \ghntwo~ \cite{knyazev2021parameter} which previously often fell short to improve over random initializations\footnote{We refer to the fine-tuning results reported by \citet{knyazev2021parameter} in Appendix E and recently in \cite{czyzewski2022breaking}.}.
The GHN-3 we introduce predicts parameters of a significantly better quality (Fig.~\ref{fig:resnet50-conv1}). We make the following \textbf{contributions}:
\vspace{-5pt}
\begin{enumerate}
 \setlength\itemsep{1pt}
 
    \item We adopt Transformer from~\citet{ying2021transformers} to improve the efficiency and scalability of GHNs and we modify it to better capture local and global graph structure of neural architectures (Section~\ref{sec:methods});
    
    \item We significantly scale up our Transformer-based GHN (GHN-3) and we release our largest model achieving the best results
    at \textcolor{violet}{\ghnurl} (Fig.~\ref{fig:scaleup-teaser});
    
    \item We evaluate GHNs by predicting parameters for around 1000 unseen ImageNet models, including all models available in the official PyTorch framework~\cite{paszke2019pytorch}. Despite such a challenging setting, our GHN-3 shows high quality and performance  and can significantly improve training of neural networks on ImageNet and other vision tasks (Section~\ref{sec:exper}).
    
\end{enumerate}

\section{Related Work}
\label{sec:related}

\textbf{Parameter prediction.}
Parameter prediction or generation is often done by hypernetworks~\cite{ha2016hypernetworks}.
Originally, hypernetworks were able to generate model parameters only for a specific architecture and dataset. Several works extended them to generalize to unseen architectures~\cite{brock2017smash,zhang2018graph} and datasets~\cite{requeima2019fast,zhmoginov2022hypertransformer}.
We focus on the unseen architectures regime where the most performant and flexible method so far is graph hypernetworks (GHNs)~\cite{zhang2018graph}. \citet{knyazev2021parameter} improved GHNs by introducing \ghntwo predicting parameters for unseen architectures with relatively high performance in image classification tasks. To train and evaluate GHNs, they introduced a diverse and large dataset of training and evaluation architectures -- \dataset. Our proposed \ghnthree closely resembles \ghntwo and uses the same training dataset. However, \ghnthree is $>100\times$ larger, which we show is important to increase the performance on ImageNet. At the same time, \ghnthree is efficient during training (and comparable during inference, see Fig.~\ref{fig:resnet50-conv1}) due to using transformer layers as opposed to a slow \mbox{GatedGNN} of \ghntwo that required sequential graph traversal (Section~\ref{sec:background}).

\textbf{Generative models of neural networks.}
Another line of work is based on first collecting a dataset of trained networks and then learning a generative model by fitting the distribution of trained weights.
\citet{schurholt2022hyper} train an auto-encoder with a bottleneck representation with the ability to sample new weights.
\citet{peebles2022learning} train a generative diffusion transformer to generate weights conditioned on random initialization for the same architecture and dataset.  
\citet{ashkenazi2022nern} improve the quality of generated parameters by combining hypernetworks and generative models, but their model is architecture-specific and unable to generate weights for unseen architectures as GHNs.\looseness-1

\textbf{Data-driven initialization.}
GHNs can be viewed as a data-driven initialization method. Predicted parameters are generally inferior than those trained with SGD for many epochs, so using the predicted ones as initialization (Eq.~\ref{eq:predict}) followed by fine-tuning with SGD (Eq.~\ref{eq:finetune}) is a logical approach.
GradInit~\cite{zhu2021gradinit} and
MetaInit~\cite{dauphin2019metainit} are methods to initialize a given neural network on a given dataset by adjusting the weights to improve the gradient flow properties.
Neural initialization
optimization (NIO)~\cite{yang2022towards} further improves on them.
These methods generally outperform a standard random-based initialization~\cite{he2015delving} and carefully designed architecture-specific initialization rules~\cite{zhang2019fixup}. 
Compared to these, in our work the initial parameters are predicted by a GHN given a computational graph of the neural network. Our approach leverages a million of training architectures making the predicted parameters better for initialization as we show.
Another recent initialization method~\cite{czyzewski2022breaking} requires a source network and the quality of the initialization depends on how similar are the source and target networks. Their work is related to Net2Net~\cite{chen2015net2net} and GradMax~\cite{evci2022gradmax} that require a smaller variant of the network or a certain growing scheduler. Our GHN-based approach is more flexible as we can initialize a larger variety of networks without the need of source networks or growing schedules.
Additional related work is also discussed in Section~\ref{apdx:related-work-nas}.

\section{Background}
\label{sec:background}

\subsection{Graph HyperNetworks}
\label{sec:bg_ghn}

\setlength{\belowdisplayskip}{1pt}
\setlength{\abovedisplayskip}{1pt}

A Graph HyperNetwork (GHN)~\cite{zhang2018graph,knyazev2021parameter} is a neural network $H_\domain$ parameterized by $\theta$ and trained on a dataset $\domain$. The input to the GHN is a computational graph $f^G$ of a neural network $f$; the output is its parameters $\w_\text{pred}$: $\w_\text{pred} = H_\domain(f^G;{\theta})$.
In our context, $\domain$ can be an ImageNet classification task, $f$ can be ResNet-50 while $\w_\text{pred}$ are parameters of ResNet-50's convolutional, batch normalization and classification layers.
\citet{knyazev2021parameter} train the GHN $H_D$ by running SGD on the following optimization problem over $M$ training architectures $\{f^G_a\}_{a=1}^M$ and $N$  training samples $\{\bx_j, y_j\}_{j=1}^N$:

\begin{equation}
 \label{eq:solution}
     \underset{\theta}{\text{arg\,min }} \frac{1}{NM}\sum_{j=1}^N \sum_{a=1}^{M}\mathcal{L}\Big(f_a\Big( \bx_j; H_\domain(f^G_a;{\theta})\Big), y_j\Big).
\end{equation}
During training $H_D$, a meta-batch of $m$ training architectures is sampled for which $H_D$ predicts parameters. Simultaneously, a mini-batch of $b$ training samples $\bx$ is sampled and forward propagated through the predicted parameters for $m$ architectures to predict $m \times b$ sample labels $\hat{y}$. For classification, the cross-entropy loss $\mathcal{L}$ is computed between $\hat{y}$ and ground truth labels $y$ of $\bx$, after which the loss is backpropagated to update the parameters $\theta$ of $H_D$ with gradient descent.
In GHN-2 and in this work, training architectures are sampled from DeepNets-1M -- a dataset of 1 million architectures~\cite{knyazev2021parameter}.
Training samples $\bx$ represent images and $y$ are their labels.

The input computational graph $f^G=(V,E)$ is a directed acyclic graph (DAG), where the nodes $V$ correspond to operations (convolution, pooling, self-attention, etc.), while the edges $E$ correspond to the forward pass flow of inputs through the network $f$. 
GHNs take $d$-dimensional node features $\mathbf{H}^{(1)} \in \mathbb{R}^{|V| \times d}$ as input obtained using an embedding layer for each $i$-th node: $\h^{(1)}_i = \text{Embed}(\h^{(0)}_i)$, where $\h^{(0)}_i$ is a one-hot vector denoting an operation (e.g. convolution).
The graph $f^G$ is traversed in the forward and backward directions and features $\mathbf{H}^{(1)}$ are sequentially updated using a gated graph neural network (GatedGNN)~\cite{li2015gated}.
After the GatedGNN updates node features, the decoder uses them to predict the parameter tensor of the predefined shape. In \ghntwo, this shape is equal to $2d \PLH 2d \PLH 16 \PLH 16$. The final predicted parameters $\w_\text{pred}$ associated with each node are obtained by copying or slicing this tensor as necessary.

\subsection{Transformers}

Transformers are neural networks with a series of self-attention (SA)-based layers applied to $d$-dimensional features $\mathbf{H} \in \mathbb{R}^{n \times d}$~\cite{vaswani2017attention}.
In a Transformer layer, SA projects $\mathbf{H}$ using learnable parameters $\mathbf{W}_Q, \mathbf{W}_K, \mathbf{W}_V$ followed by the pairwise dot product, softmax and another dot product (for simplicity we omit the layer superscript $^{(l)}$ in our notation unless necessary):
\begin{align}
    \label{eq:sa_proj}
    Q &= \mathbf{H}\mathbf{W}_Q, K = \mathbf{H}\mathbf{W}_K, V =\ \mathbf{H}\mathbf{W}_V, \\
    \label{eq:sa_base}
    \mathbf{A} &= \frac{\mathbf{Q}\mathbf{K}^T}{\sqrt{d}}, \ \ \text{SA}(\mathbf{H}) =\ \text{softmax}(\mathbf{A}) \mathbf{V}.
\end{align}
A Transformer layer consists of multi-head SA (MSA)
with $k$ heads, and a series of fully-connected and normalization layers~\cite{vaswani2017attention,dosovitskiy2020image}.

\subsection{Transformers on Graphs}

A vanilla Transformer defined above can in principle be applied to graph features $\mathbf{H}^{(1)} \in \mathbb{R}^{|V| \times d}$, but it does not have ingredients to capture the graph structure (edges).
Several variants of Transformer layers on graphs have been proposed~\cite{dwivedi2020generalization,ying2021transformers,kim2022pure,chen2022structure}.
We rely on Graphormers~\cite{ying2021transformers} due to its simplicity and strong ability to capture local and global graph structure.
In Graphormers, node features are augmented with node degree (centrality) embeddings. For directed graphs, such as DAGs, these are in-degree $\bz_{\text{deg}^+(i)}$ and out-degree embeddings $\bz_{\text{deg}^-(i)}$, so an embedding $\h^{(1)}_i$ for the $i$-th node is defined as: 
\begin{equation}
    \label{eq:embed_base}
    \h^{(1)}_i = \text{Embed}(\h^{(0)}_i) + \bz_{\text{deg}^+(i)} + \bz_{\text{deg}^-(i)},
\end{equation}
where $\text{Embed}(\h^{(0)}_i)$ are some initial node embeddings, e.g. see in Section~\ref{sec:bg_ghn}.
When self-attention in Eq.~\eqref{eq:sa_base} is applied to graphs, we can interpret $\mathbf{A}_{ij}$ as a scalar encoding the relationship between nodes $i$ and $j$. 
To incorporate the graph structure in Transformers, Graphormer layers add an edge embedding $\mathbf{e}(i,j)$ to $\mathbf{A}_{ij}$ in self-attention:
\begin{equation}
    \label{eq:sa_graphormer}
    \mathbf{\tilde{A}}_{ij} = \frac{(\h_i\mathbf{W}_Q) (\h_j\mathbf{W}_K)^T}{\sqrt{d}} + \mathbf{e}(i,j),
\end{equation}
where $\mathbf{\tilde{A}}$ replaces  $\mathbf{A}$ in Eq.~\eqref{eq:sa_base} without changing the other components of the Transformer layer.
Graphormers model $\mathbf{e}(i,j)$ as a learnable scalar dependant on the shortest path distance between nodes $i$ and $j$, shared across all Graphormer layers. While for SA $\mathbf{e}(i,j)$ is a scalar, for MSA $\mathbf{e}(i,j)$ is a vector of length equal to the number of heads $k$. See further details in~\cite{ying2021transformers}.

\section{Scalable Graph HyperNetworks: \ghnthree}
\label{sec:methods}

Our \ghnthree model modifies \ghntwo~\cite{knyazev2021parameter} in three key ways: (1) replacing a GatedGNN with Graphormer-based layers; (2) improving the training loss of GHNs; (3) increasing the scale and meta-batch of GHNs.\looseness-1

\subsection{Transformer on Computational Graphs}

In \ghnthree, we replace a GatedGNN with a stack of $L$ Graphormer-based layers built based on Eq.~\eqref{eq:sa_graphormer} and applied to $\mathbf{H}^{(1)}$ defined in Eq.~\eqref{eq:embed_base}. 
Since the information in computational graphs can flow in the forward ($i \rightarrow j$) and backward ($j \rightarrow i$) directions, we explicitly separate edge embeddings into two terms, so that our self-attention is based on:\looseness-1
\begin{equation}
\label{eq:sa}
    \mathbf{\tilde{A}}_{ij} = \frac{(\h_i\mathbf{W}_Q) (\h_j\mathbf{W}_K)^T}{\sqrt{d}} + \phi([\mathbf{e}(i,j), \mathbf{e}(j,i)]).
\end{equation}
where $\phi$ is a series of fully connected layers.
While in principle multiple Graphormer layers (Eq.~\ref{eq:sa_graphormer}) could infer that the backward pass is the inverse of the forward pass, we found that explicitly separating the embeddings and adding $\phi$ facilitates learning as we confirm in Ablations (Section~\ref{sec:exper}).

\subsection{Predicted Parameter Regularization}
\label{sec:wd}

\citet{knyazev2021parameter} showed that the parameters predicted by GHNs can lead to a higher variance of activations compared to the variances of activations for parameters optimized with SGD (see their Appendix B). Too high variances may lead to instabilities during training GHNs as well as to negative effects in applications. For example, if such predicted parameters are used for initialization, their fine-tuning can be challenging. To alleviate this issue, we add a group $\ell_1 \!-\! \ell_2$ regularization~\cite{scardapane2017group} on the  predicted parameters $\w_\text{pred}$ during training GHN-3, so that the total \mbox{GHN-3} training loss becomes:
\begin{equation}
\label{eq:wd}
    \mathcal{L} = \mathcal{L}_{CE} + \gamma \sum\nolimits_i || \w_{\text{pred},i} ||_2,
\end{equation}
where $\mathcal{L}_{CE}$ is the cross-entropy loss (same as in the baseline \ghntwo), $\gamma$ is a tunable coefficient controlling the strength of the penalty and $i$ is the layer index of the training neural network $f_a$ (see Eq.~\ref{eq:solution}).
Our regularization encourages smaller values in predicted parameters and consequently smaller variances of activations that are expected to be more aligned with the activations of models trained with SGD (see Fig.~\ref{fig:val_curves}-bottom).
Besides Eq.~\eqref{eq:wd}, we also experimented with other forms of regularization including $\sum\nolimits_i  \w_{\text{pred},i}^2$, but Eq.~\eqref{eq:wd} overall worked the best (see ablations in Table~\ref{tab:ablations}).

\subsection{Increased Scale} Once we replace GatedGNN with Graphormer-based layers, scaling up GHNs becomes straightforward by increasing the number of the Graphormer-based layers $L$ and hidden size $d$. 
The decoder of our \ghnthree has the same architecture as in \ghntwo with the hidden size increasing proportionally to $d$ (see Appendix~\ref{apdx:details}). The decoder takes the output node features 
of the last Graphormer layer to predict parameters.
In contrast to our approach, scaling up \ghntwo is not trivial since a single GatedGNN layer is computationally expensive as it requires sequential graph traversal, so stacking such layers is not feasible (see Section~\ref{sec:exper}). Instead, Graphormer layers update all node features in parallel making our \ghnthree efficient and scalable. Increasing the meta-batch size $m$ (see Section~\ref{sec:bg_ghn}) is also straightforward, so when training a GHN on $g$ GPU devices, each device processes $m/g$ architectures and the gradients of the GHN parameters $\theta$ are averaged on the main device~\cite{knyazev2021parameter}.\looseness-1

Despite the simple modifications outlined in this Section, \ghnthree brings GHNs to a significantly better level in the quality of predicted parameters with important practical implications as we empirically show next.

\section{Experiments}
\label{sec:exper}

We evaluate if neural networks initialized with the parameters $\w_\text{pred}$ predicted by GHNs obtain high performance without any training (Eq.~\ref{eq:predict}) and after fine-tuning (Eq.~\ref{eq:finetune}). 
We focus on a large-scale ImageNet setting, but also evaluate in a transfer learning setting from ImageNet to few-shot image classification and object detection tasks.

\textbf{Training GHN-3.}
We train the GHNs on the ILSVRC-2012 ImageNet dataset~\cite{russakovsky2015imagenet} with 1.28M training and 50K validation images of the 1k classes.
All GHNs are trained for 75 epochs using AdamW~\cite{loshchilov2017decoupled}, initial learning rate \mbox{4e-4} decayed using the cosine scheduling, weight decay $\lambda$=\mbox{1e-2}, predicted parameter regularization $\gamma$=3e-5 (Eq.~\ref{eq:wd}), batch size $b$=128 and automatic mixed precision in PyTorch~\cite{paszke2019pytorch}.
We train the GHNs on the same training split of 1 million architectures, \dataset, used to train \ghntwo~\cite{knyazev2021parameter}.\looseness-1

\textbf{\ghnthree variants.}
We train \ghnthree models of four scales (Table~\ref{tab:models}). We begin with a tiny scale \ghnt that has the same order of parameters (6.9M) as \ghntwo (2.3M). 
We then gradually increase the number of layers, hidden size and heads in the Graphormer-based layers following a common style in Transformers~\cite{dosovitskiy2020image,liu2021swin}.
We train all variants with meta-batch size $m$=8 and 16 and denote the GHNs with the /m8 and /m16 suffixes.\looseness-1

\begin{figure}[t]
\begin{center}
\centerline{\includegraphics[width=0.95\columnwidth]{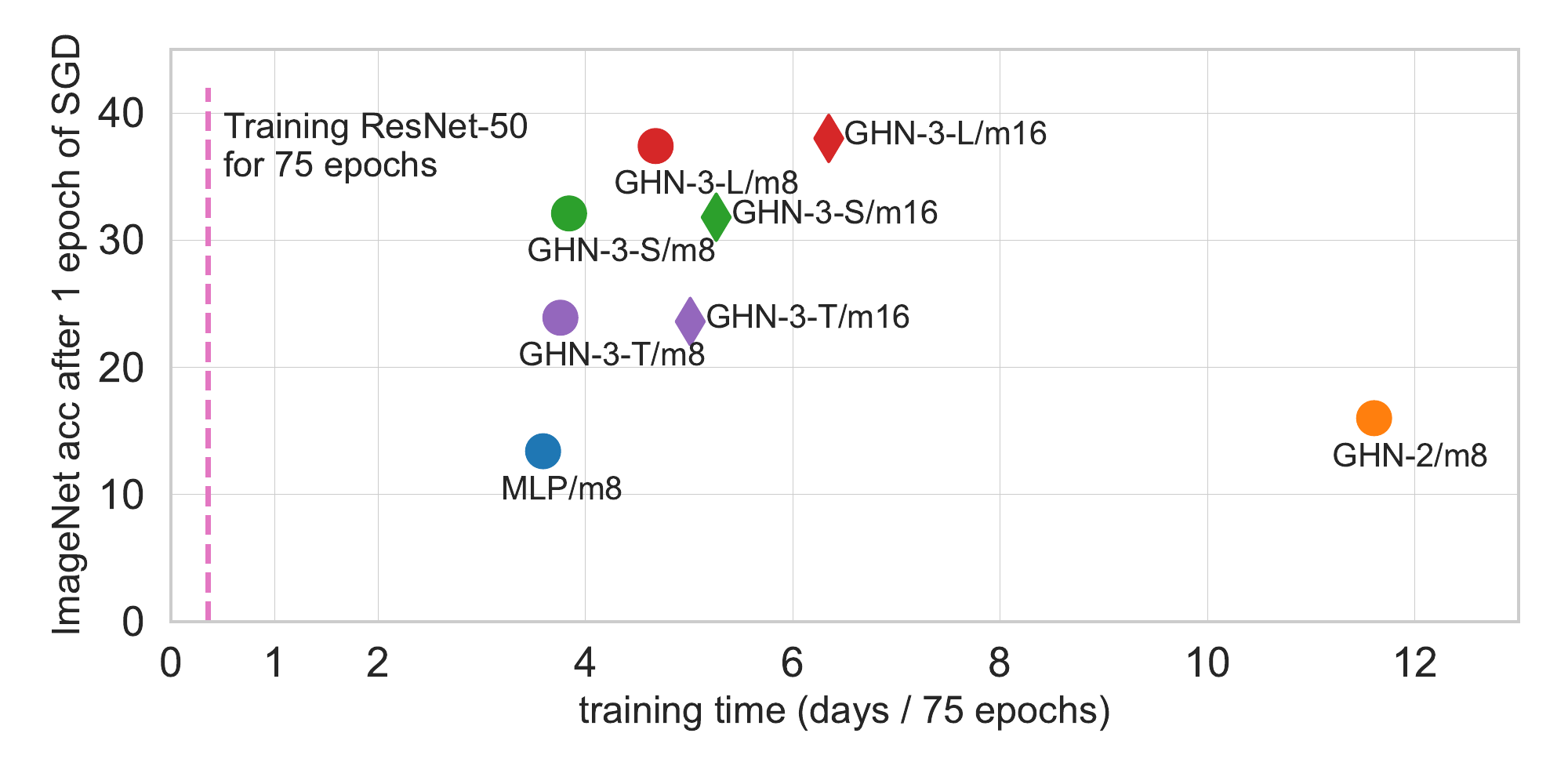}}
\vspace{-14pt}
\caption{ImageNet accuracy of fine-tuning predicted parameters \textit{vs} training speed of GHNs. Our \ghnthree is efficient to train and at the same time the predicted parameters achieve higher accuracy. All measurements are made on 4xNVIDIA-A100 GPUs.
}
\label{fig:acc_speed}
\end{center}
\vskip -0.3in
\end{figure}

\textbf{Efficient distributed implementation.}
We build on the open-source implementation of \ghntwo~\cite{knyazev2021parameter} and improve its efficiency by using a distributed training pipeline and removing redundant computations.
Our implementation reduces the training time of GHNs by around 50\% (see Appendix~\ref{apdx:results}). However, despite our best efforts GHN-2 still takes more than 11 days to train (Fig.~\ref{fig:acc_speed}). Larger versions of GHN-2, e.g. by stacking two GatedGNN layers, are estimated to require $>$ 20 days making it expensive to train. In contrast, our GHN-3 is significantly faster to train while achieving stronger results (Fig.~\ref{fig:acc_speed}). For example, one of our best performing models (\ghnl/m16) is about $2\PLH$ faster to  train and is more than 2$\PLH$ better in accuracy.
The cost of training our GHNs is still higher than training a single network (e.g. full training of ResNet-50 takes about 0.4 days on 4xNVIDIA-A100). However, we train each GHN only once and we publicly release them, so that they can predict parameters for many diverse and large ImageNet models in $\approx 1$~second even on a CPU (Fig.~\ref{fig:resnet50-conv1}).\looseness-1

\begin{table}[t]
    \centering
    \caption{Details of GHNs. Train time is for GHNs with $m=8$ and is measured on 4xNVIDIA-A100 GPUs. \textsuperscript{*}\ghnxl requires 8 GPUs, so its training time is not directly comparable. \textcolor{gray}{\ghntwo-S} is a larger version of the baseline GHN-2 that we attempted to train but were unable to complete due to poor training efficiency.}
    \label{tab:models}
    \vskip 0.05in
    \begin{center}
    \begin{scriptsize}
    \begin{sc}
    \setlength{\tabcolsep}{2pt}
    \begin{tabular}{lccccc}
    \toprule
    \textbf{Name} & \textbf{Layers} $L$ & \textbf{Hidden size} $d$ & \textbf{Heads} $k$ & \textbf{Params} & \textbf{Train time}\\
    \midrule
    \ghntwo & 1 & 32 & $-$ & 2.3M & 11.5 days\\
    \textcolor{gray}{\ghntwo-S} & \textcolor{gray}{2} & \textcolor{gray}{128} & \textcolor{gray}{$-$} & \textcolor{gray}{35.0M} & \textcolor{gray}{20.5 days} \\  \midrule
    \ghnt & 3 & 64 & 8 & 6.9M & 3.7 days \\
    
    \ghns & 5 & 128 & 16 & 35.8M & 3.8 days \\
    
    \ghnl & 12 & 256 & 16  & 214.7M & 4.7 days \\
     
    \ghnxl & 24 & 384 & 16 & 654.4M & 4.8\textsuperscript{*} days \\
    
    \bottomrule
    \end{tabular}
    \end{sc}
    \end{scriptsize}
    \end{center}
    \vskip -0.15in
\end{table}

\textbf{Baselines.} Our main baselines are \ghntwo, released by~\citet{knyazev2021parameter}, random initialization (\rand) implemented by default in PyTorch, typically based on~\citet{he2015delving}. In full training experiments, we also compare our approach to more advanced initializations, \grad~\cite{zhu2021gradinit} and NIO~\cite{yang2022towards}.
For reference, we also compare to the \textsc{SOTA} results reported in prior literature. However, this baseline is not a fair comparison since it requires 90-600 epochs of training on ImageNet and often relies on advanced optimization algorithms, augmentations and other numerous tricks.\looseness-1

\textbf{Evaluation architectures.}
For evaluation of GHNs we use two splits of neural network architectures. 
(1) The evaluation splits of \dataset: 900 in-distribution (Test split) and out-of-distribution (Wide, Deep, Dense and BN-Free splits) architectures~\cite{knyazev2021parameter}.
(2) \textsc{PyTorch}: 74 architectures for which trained ImageNet weights are available in PyTorch-v1.13~\cite{paszke2019pytorch}.
As shown by~\citet{knyazev2021parameter}, all the evaluation architectures of \dataset are different from the ones used to train GHNs. The architectures of \textsc{PyTorch} are even more different, diverse and on average of a significantly larger scale than the ones in \dataset (Table~\ref{tab:nets}).

\begin{table*}[t]
    \vspace{-7pt}
    \centering
    \caption{ImageNet top-1 accuracy (\%) after predicting parameters and fine-tuning them for 1/10 (1k SGD steps) and 1 (10k steps) epoch. Average and standard deviation of accuracies for top-10 networks in the \dataset and \textsc{PyTorch} splits are reported. \textcolor{gray}{SOTA} results denote accuracies reported in the official PyTorch documentation and require 90-600 epochs, advanced optimizers and numerous tricks.\looseness-1
    }
    \label{tab:bench_imagenet}
    \vskip 0.05in
    \begin{center}
    \begin{scriptsize}
    \begin{sc}
    \setlength{\tabcolsep}{5pt}
    \begin{tabular}{l|cccccc|c}
        \toprule
        \textbf{Initialization} &
        \multicolumn{2}{c}{\textbf{No fine-tuning}} & \multicolumn{2}{c}{\textbf{1/10 epoch of SGD}} & \multicolumn{2}{c|}{\textbf{1 epoch of SGD}} & 
        \textcolor{gray}{SOTA}\\
         & deepnets1-1m-10 & pytorch-10 & deepnets1-1m-10 & pytorch-10 & deepnets1-1m-10 & pytorch-10 & \textcolor{gray}{pytorch-10} \\
        \midrule
        
        \rand & 0.2\std{0.0} & 0.2\std{0.0} & 5.2\std{0.7} & 1.1\std{0.3} & 34.7\std{1.1} & 25.9\std{1.3} & \textcolor{gray}{84.2\std{0.7}} \\
        
        \ghntwo/m8 & 24.6\std{0.4} & 0.7\std{0.4} & 15.9\std{1.3} & 7.6\std{2.7} & 24.9\std{1.3} & 16.0\std{3.2} & 
        \multirow{9}{*}{\rotatebox[origin=c]{90}{\parbox{2cm}{\scriptsize\textcolor{gray}{90-600 epochs}}}}\\
        
        \ghnt/m8 & 31.3\std{0.3} & 3.9\std{3.9} & 31.9\std{1.9} & 15.3\std{3.4} & 33.6\std{1.5} & 23.9\std{1.0}\Tstrut\\
        
        \ghnt/m16 & 30.0\std{0.2} & 5.3\std{4.5} & 31.6\std{1.1} & 16.1\std{3.8} & 32.8\std{1.0} & 23.6\std{1.6}\\
        
        \ghns/m8 & 41.1\std{0.7} & 6.0\std{6.3} & 42.1\std{1.8} & 22.5\std{5.0} & 43.4\std{1.5} & 32.1\std{1.4}\Tstrut\\
        
        \ghns/m16 & 41.3\std{0.6} & 7.8\std{7.0} & 41.9\std{2.5} & 22.0\std{5.9} & 43.5\std{1.7} & 31.8\std{2.0}\\

        \ghnl/m8 & 43.6\std{1.0} & 10.7\std{9.3} & 44.0\std{2.8} & 26.2\std{5.1} & 46.0\std{2.0} & 37.4\std{1.9}\Tstrut\\ 
        
        \ghnl/m16 & 44.7\std{0.7} & 9.8\std{8.1} & 42.9\std{5.6} & 27.3\std{4.6} & 46.4\std{3.1} & 38.0\std{2.3} \\

        \ghnxl/m8 & 43.8\std{0.8} & 8.5\std{8.1} & 45.7\std{2.5} & 28.2\std{6.1} & 47.5\std{2.3} & 36.9\std{3.8}\Tstrut\\

        \ghnxl/m16 & \textbf{47.1}\std{0.9} & \textbf{11.5}\std{8.4} & \textbf{48.1}\std{3.7} & \textbf{30.6}\std{7.2} & \textbf{50.1}\std{3.0} & \textbf{40.3}\std{3.9}\\
        
         \bottomrule
    \end{tabular}
    \end{sc}
    \end{scriptsize}
    \end{center}
    \vskip -0.2in
\end{table*}

\begin{figure*}[h!]
\begin{center}
\setlength{\tabcolsep}{1pt}
\begin{tabular}{cc}
     \includegraphics[width=0.75\textwidth]{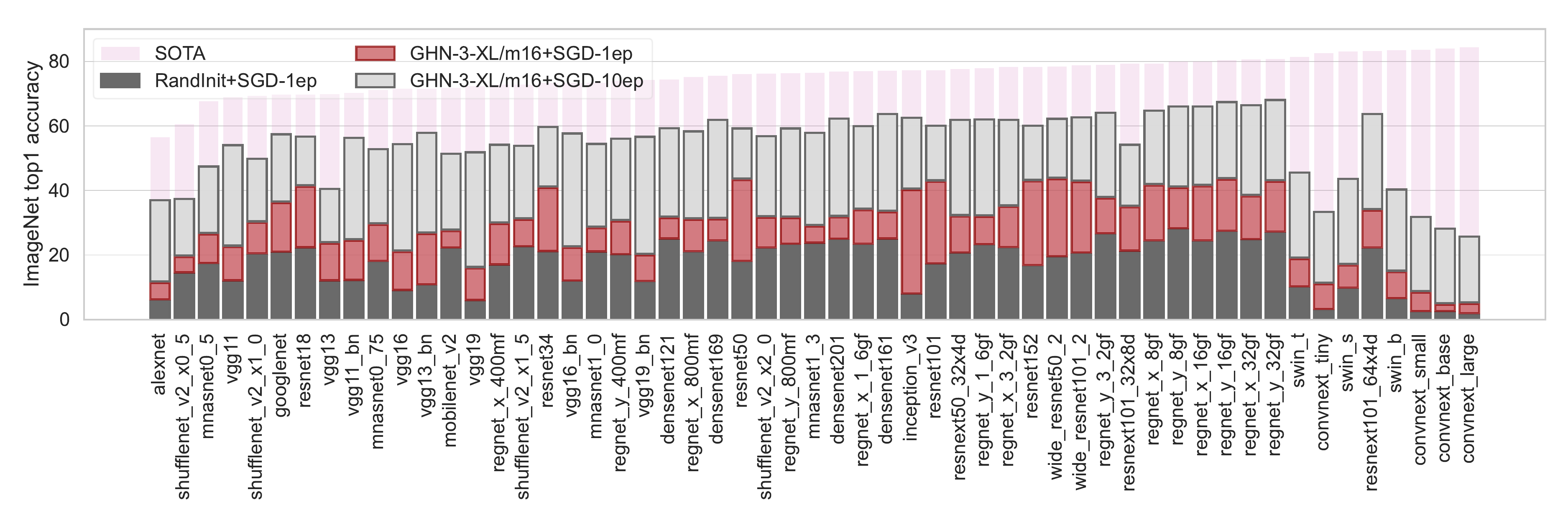} & \includegraphics[width=0.25\textwidth]{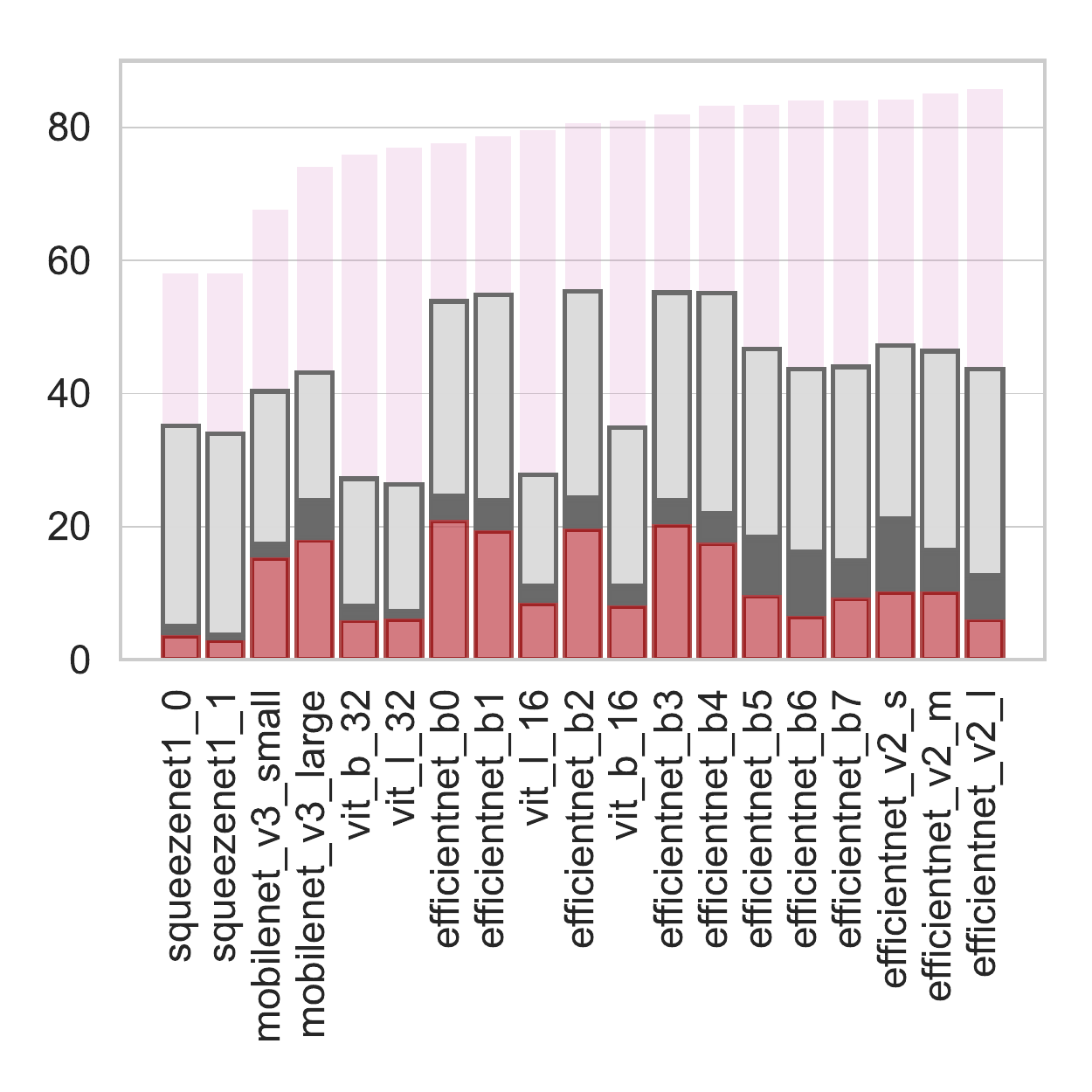} \\
\end{tabular}
\vspace{-14pt}
\caption{ImageNet results for all \textsc{PyTorch} models after 1 and 10 epochs of SGD, and after SOTA training for at least 90 epochs. 
The \textbf{left} and \textbf{right} figures show the models for which \ghnxl/m16+SGD-1ep \textbf{outperforms} and \textbf{underperforms} \randnospace+SGD-1ep. See accuracies and rank correlation results in Table~\ref{tab:ghn3-full-results}. 
}
\label{fig:pytorch}
\end{center}
\vskip -0.2in
\end{figure*}

\begin{table}[t]
    \centering
    \vspace{-7pt}
    \caption{The evaluation splits of neural architectures. The average, standard deviation and range of property values are shown. Graph degrees and paths are computed based on \citet{knyazev2021parameter}.
    }
    \label{tab:nets}
    \vskip 0.05in
    \begin{center}
    \begin{scriptsize}
    \begin{sc}
    \setlength{\tabcolsep}{1pt}
    \begin{tabular}{lcccc}
    \toprule
    \textbf{Eval. split} & \textbf{ nets} & \textbf{params (M)} & \textbf{Graph degree} & \textbf{Graph path}  \\
    \midrule
    \dataset & 900 & 9\std{16} (2.2 - 102) & 2.3\std{0.1} (2.1 - 2.8) & 15\std{7} (5.1 - 72) \\

    PyTorch & 74 &  56\std{64} (1.2 - 307) & 2.5\std{1.2} (1.9 - 9.1) & 16\std{6} (6.8 - 37) \\
    
    \bottomrule
    \end{tabular}
    \end{sc}
    \end{scriptsize}
    \end{center}
    \vskip -0.2in
\end{table}

\subsection{Experiments on \dataset and \textsc{PyTorch}}

\textbf{Setup.}
Using trained GHNs we first predict ImageNet parameters for all 900 + 74 evaluation architectures in \dataset and \textsc{PyTorch} and evaluate their ImageNet classification accuracy (top-1) by propagating validation images. 
We then initialize the networks (1) with parameters predicted by GHNs or (2) randomly (\rand), and fine-tune them using SGD on ImageNet to compare convergence dynamics between (1) and (2). 
Since fine-tuning all the 974 networks for all initialization approaches is prohibitive, we choose top-10 networks in each split, denoted as \dataset-10 and \textsc{PyTorch}-10 respectively.
For the GHNs, the top-10 networks are chosen based on the accuracy obtained by directly evaluating (no fine-tuning) predicted parameters on ImageNet.
For \rand, the top-10 networks are chosen based on the network accuracy after 1 epoch of SGD (denoted as SGD-1ep). For \dataset, the accuracies of \randnospace+SGD-1ep are taken from~\cite{knyazev2021parameter}. For \textsc{PyTorch} we trained all networks for 1 epoch using SGD with a range of learning rates (0.4, 0.1, 0.04, 0.01, 0.001), momentum 0.9, weight decay 3e-5 and batch size 128. 
We choose the networks achieving the best validation accuracy among all the learning rates and report the average accuracy (Table~\ref{tab:bench_imagenet}).
For the GHNs, we fine-tune top-10 networks initialized with predicted parameters using SGD for 1 epoch using the same range of hyperparameters as for training from \rand.
For our best \ghnxl/m16 we also fine-tune all \textsc{PyTorch} networks.\looseness-1

\begin{table}[t]
    \centering
    \vspace{-7pt}
    \caption{Summary of the results presented in Fig.~\ref{fig:pytorch}.
    ``Wins'' denotes a fraction of networks for which a GHN-3-based init. outperformed \randnospace+SGD-1ep.
    ``Avg gain/loss'' denotes an average gain/loss versus \randnospace+SGD-1ep when GHN-3 wins/loses.
    }
    \label{tab:wins}
    \vskip 0.05in
    \begin{center}
    \begin{small}
    \begin{sc}
    \setlength{\tabcolsep}{4pt}
    \begin{tabular}{lccc}
        \toprule
        \textbf{Method} & \textbf{Wins} & \textbf{Avg gain} & \textbf{Avg loss} \\
        \midrule
        \scriptsize \ghnxl/m16 no fine-tune & 5\% & 1.6\% & -16.4\% \\
        \scriptsize \ghnxl/m16  + SGD-1/10ep & 14\% & 11.0\% & -12.0\% \\
        \scriptsize \ghnxl/m16 + SGD-1ep & 74\% & 12.3\% & -4.2\%  \\
        
         \bottomrule
    \end{tabular}
    \end{sc}
    \end{small}
    \end{center}
    \vskip -0.2in
\end{table}

\textbf{Results of fine-tuning top-10 models for 1 epoch.}
As reported in Table~\ref{tab:bench_imagenet}, our GHN-3-based initialization consistently improves ImageNet performance compared to \rand and the GHN-2-based initialization for all training regimes and for both \dataset-10 and \textsc{PyTorch}-10 architecture splits.
Moreover, GHN-3 results gracefully improve with the GHN scale.
For larger GHN-3 models, increasing meta-batch size ($m$) is important. For example, when $m$ is increased from 8 to 16, ImageNet accuracy on \textsc{PyTorch}-10 after 1 epoch increases by 3.4 points for \ghnxl versus a -0.3 point decrease for \ghnt. The overall trend indicates that further increase of the GHN-3 scale and meta-batch size should yield more gains (Fig.~\ref{fig:scaleup-teaser}). 
Our best model, \ghnxl/m16, when used as an initializer for \textsc{PyTorch}-10 models leads to fast convergence. For example, after fine-tuning predicted parameters for just 1/10 epoch (1k SGD steps) we achieve 30.6\% while \rand achieves only 1.1\% after 1/10 epoch and 25.9\% after 1 epoch.
The GHN-2-based initialization is considerably worse than \rand leading to 16.0\% after 1 epoch. Thus, our GHN-3 models and in particular \ghnxl/m16 make a major step from GHN-2 by making GHNs useful for initialization.\looseness-1

\textbf{Results of fine-tuning all PyTorch models for 1-10 epochs.}
Using our best model, \ghnxl/m16, we initialized and then fine-tuned for 1 epoch all 74 \textsc{PyTorch} models and compared to \rand (Fig.~\ref{fig:pytorch}). \ghnxl/m16 improves \rand in 74\% (55 out of 74) cases with an average accuracy gain of 12.3\% in absolute points (Table~\ref{tab:wins}, Fig.~\ref{fig:pytorch}-left).
While our initialization is inferior in 26\% cases, the accuracy drop is relatively small and is -4.2\% on average (Table~\ref{tab:wins}, Fig.~\ref{fig:pytorch}-right). Most of the lost cases correspond to the networks with a squeeze and excitation operation (EfficientNet, MobileNet), indicating a potential problem of GHNs to predict good parameters in that case.
Nevertheless, when fine-tuning for 10 epochs, all \textsc{PyTorch}  models initialized using \ghnxl/m16 improve their performance fast and in some cases approach SOTA training from \rand for $\geq$90 epochs. For example, RegNet-y-32gf achieves 68\% after 10 epochs outperforming six SOTA-trained models.

\textbf{Ablations.}
We study which components of GHN-3, besides the scale and meta-batch size, are important for the performance. To do so, we apply our previous setup from Table~\ref{tab:bench_imagenet} to ablated \ghnt/m8 models (Table~\ref{tab:ablations}). 
First, we show that our predicted parameter regularization (Eq.~\ref{eq:wd}) is important.
It explicitly enforces smaller values in predicted parameters which facilitates their fine-tuning (Fig.~\ref{fig:val_curves}).
Second, adding separate edge embeddings for the backward pass of computational graphs (Eq.~\ref{eq:sa}) also improves results (from 20.4\% to 23.9\%) with almost no increase in the GHN size.
Finally, the accuracy drops the most when edge embeddings and self-attention are removed. Self-attention without edge embeddings (Eq.~\ref{eq:sa_base}) still allows for the exchange of information between the nodes of graphs, therefore the accuracy is not as low (18.9\%) as when self-attention is removed (13.4\%).
These last two results confirm the importance of capturing the graph structure of neural architectures and that our \ghnthree is an effective model to achieve that. Additional ablations are shown and discussed in Appendix~\ref{apdx:ablations}.

\begin{table}[t]
    \centering
    \vspace{-5pt}
    \caption{Ablations. ImageNet validation accuracy after 1 training epoch for the \textsc{PyTorch}-10 models initialized with \ghnt/m8.\looseness-1}
    \label{tab:ablations}
    \begin{center}
    \begin{scriptsize}
    \begin{sc}
    \setlength{\tabcolsep}{2.5pt}
    \begin{tabular}{lcccccc}
        \toprule
        \textbf{Model} & \textbf{SA} & \textbf{FW} & \textbf{BW} & $\gamma$ &  \textbf{Params}
        & \textbf{Acc}\\
        \midrule
        \ghnt/m8 & \cmark & \cmark & \cmark & 3e-5 & 6.91M & 23.9\std{1.0} \\ 
        \midrule
        
        $\gamma\sum\nolimits_i  \w_{\text{pred},i}^2$ reg. in
        Eq.~\eqref{eq:wd}  & \cmark & \cmark & \cmark & 3e-5 & 6.91M & 22.5\std{2.2} \\
        
        No pred. par. reg. in Eq.~\eqref{eq:wd}  & \cmark & \cmark & \cmark & 0 & 6.91M & 19.9\std{4.7} \\ 
        
        No bw embed. in Eq.~\eqref{eq:sa} & \cmark & \cmark & \xmark & 3e-5 & 6.90M &  20.4\std{6.9}  \\
        
        No egde embed., SA Eq.~\eqref{eq:sa_base} & \cmark & \xmark & \xmark & 3e-5 & 6.88M & 18.9\std{4.1} \\

        No egde embed., MLP & \xmark & \xmark & \xmark & 3e-5 & 6.74M & 13.4\std{0.8} \\
        \bottomrule
    \end{tabular}
    \end{sc}
    \end{scriptsize}
    \end{center}
    \vskip -0.2in
\end{table}

\begin{figure}[t]
\begin{center}
\centerline{\includegraphics[width=0.8\columnwidth]{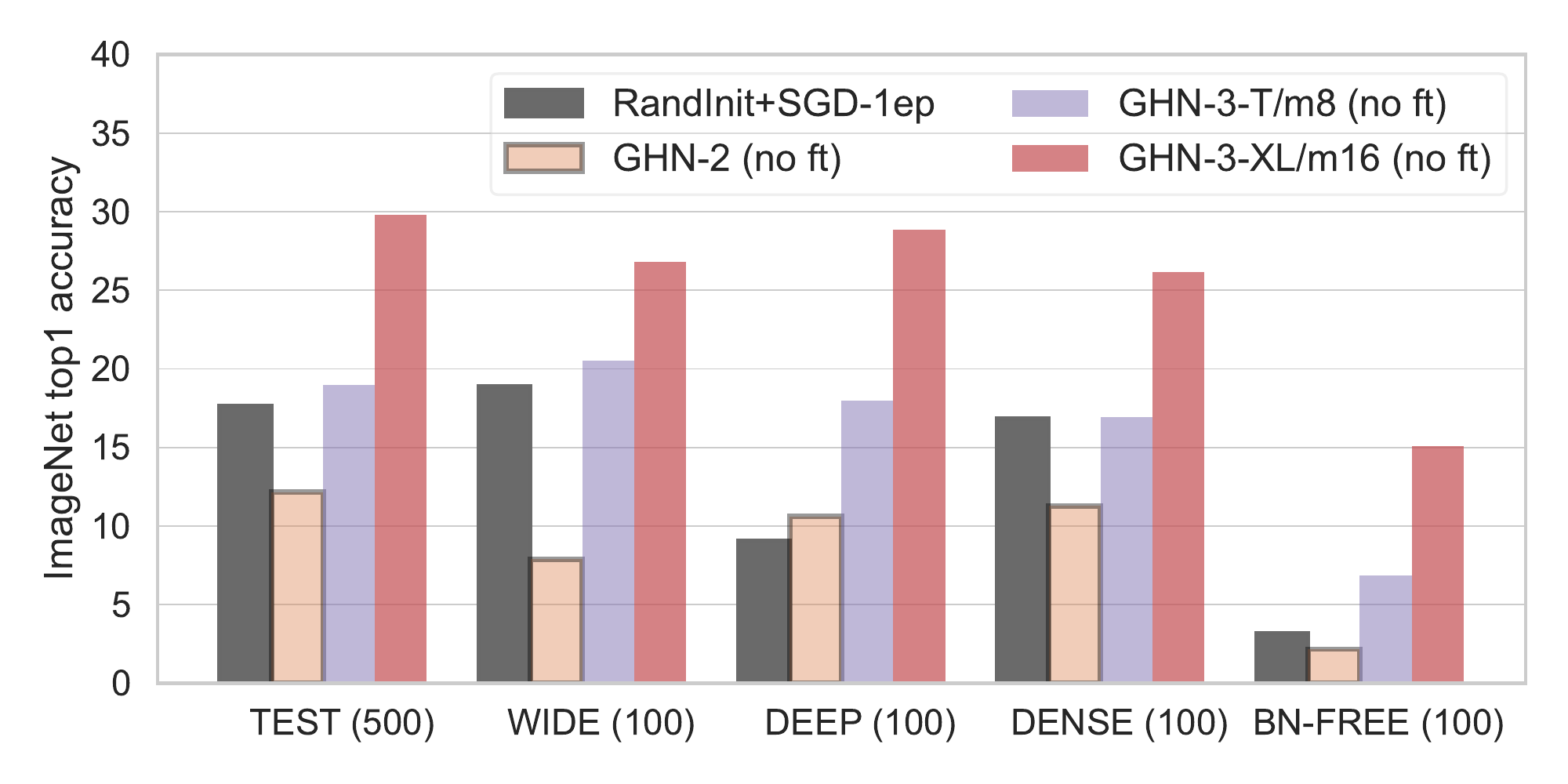}}
\vspace{-10pt}
\caption{Generalization results of GHNs \textbf{without fine-tuning} predicted parameters on five evaluation splits of \dataset using ImageNet accuracy. The number in parentheses indicates the number of architectures in each split. See the numerical results for these and other GHNs in Table~\ref{tab:generalization_more} in Appendix.
}
\label{fig:generalization}
\end{center}
\vskip -0.3in
\end{figure}

\begin{figure*}[hbpt]
\begin{center}
\setlength{\tabcolsep}{1pt}
\begin{tabular}{ccc}
    \small \textsc{ResNet-50} & \small \textsc{ResNet-152} & \small \textsc{Swin Transformer (Swin-T)} \vspace{-2pt}\\
     {\includegraphics[width=0.32\textwidth]{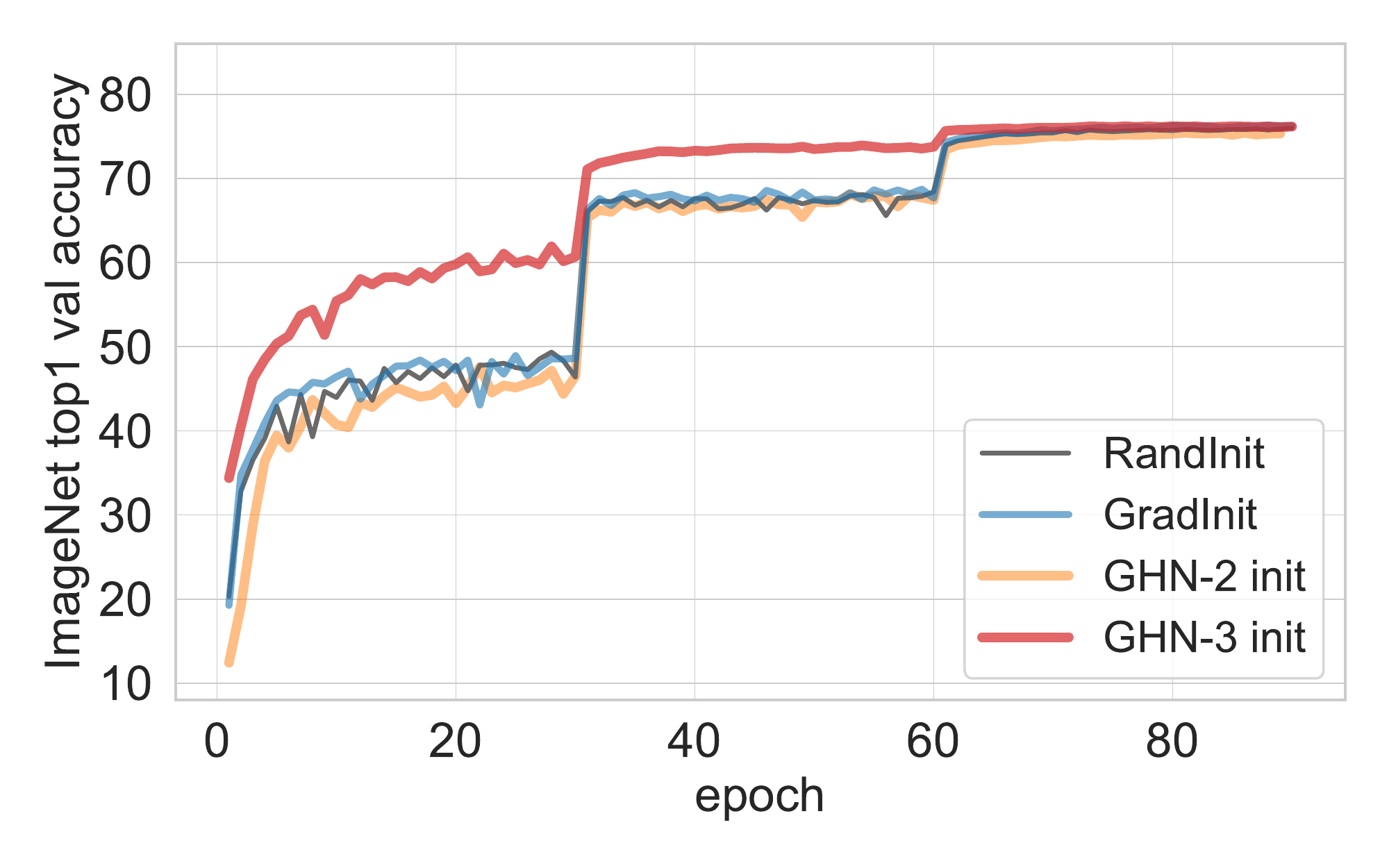}} &  {\includegraphics[width=0.32\textwidth]{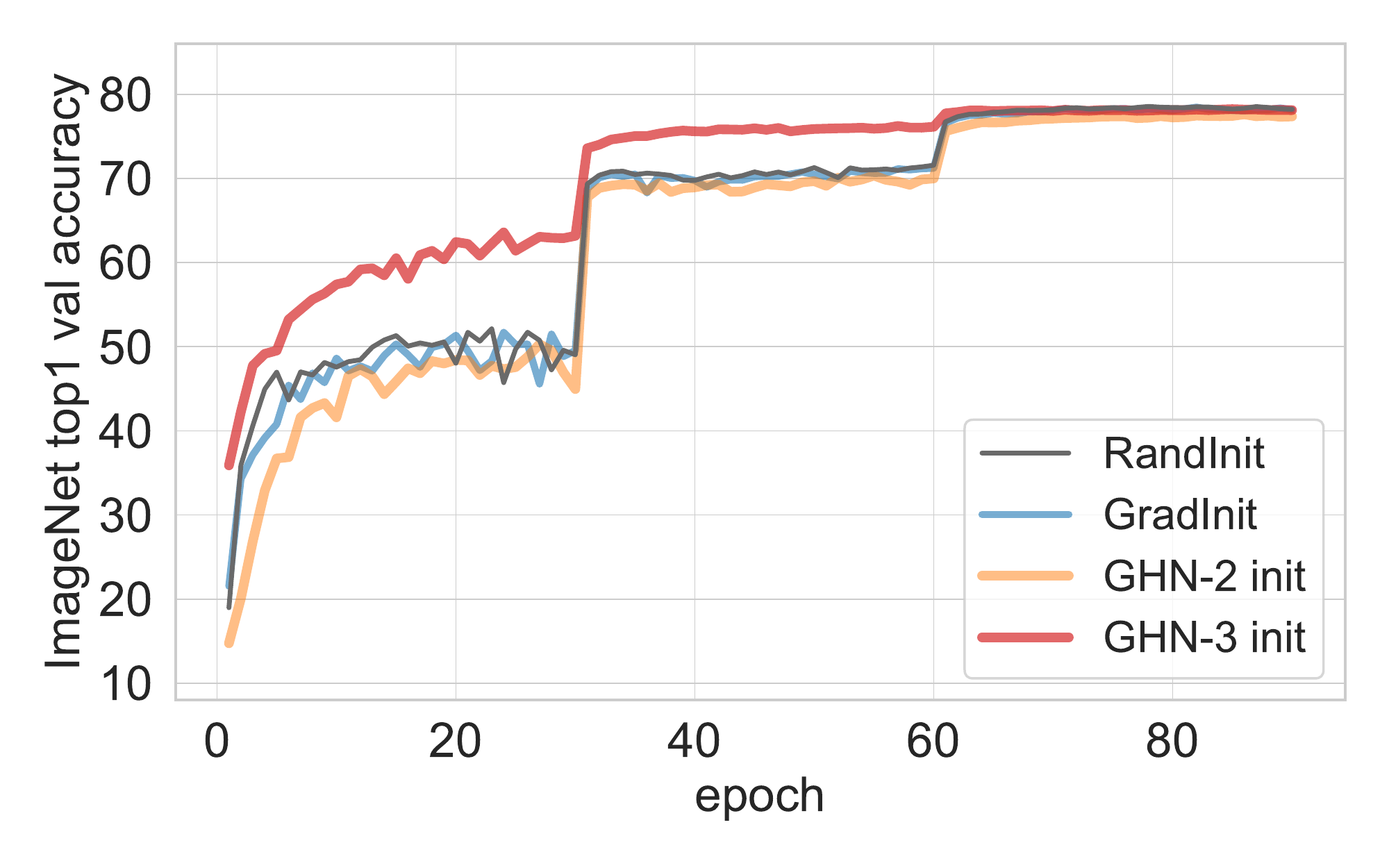}} & {\includegraphics[width=0.32\textwidth]{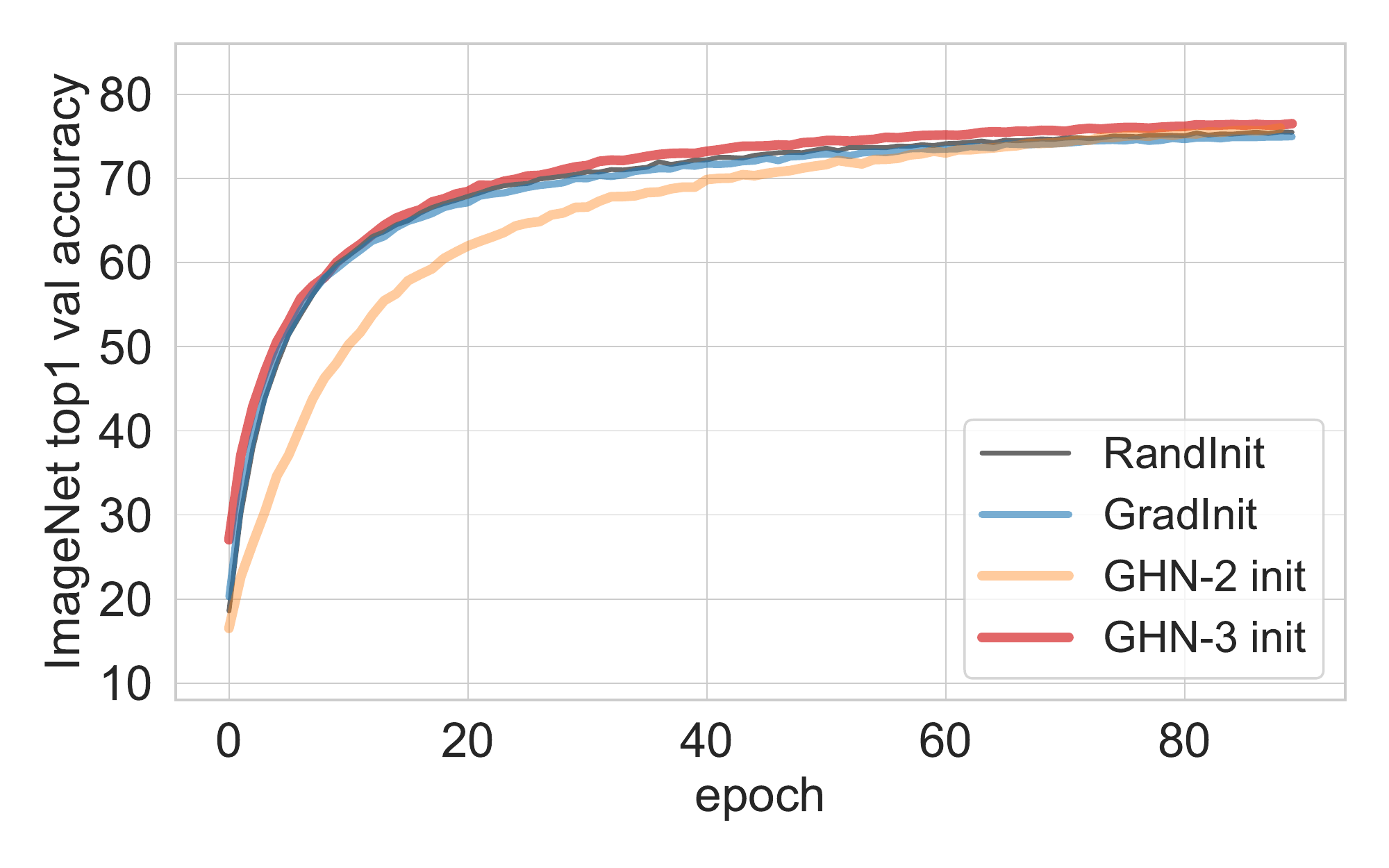}}\\
     \includegraphics[width=0.32\textwidth]{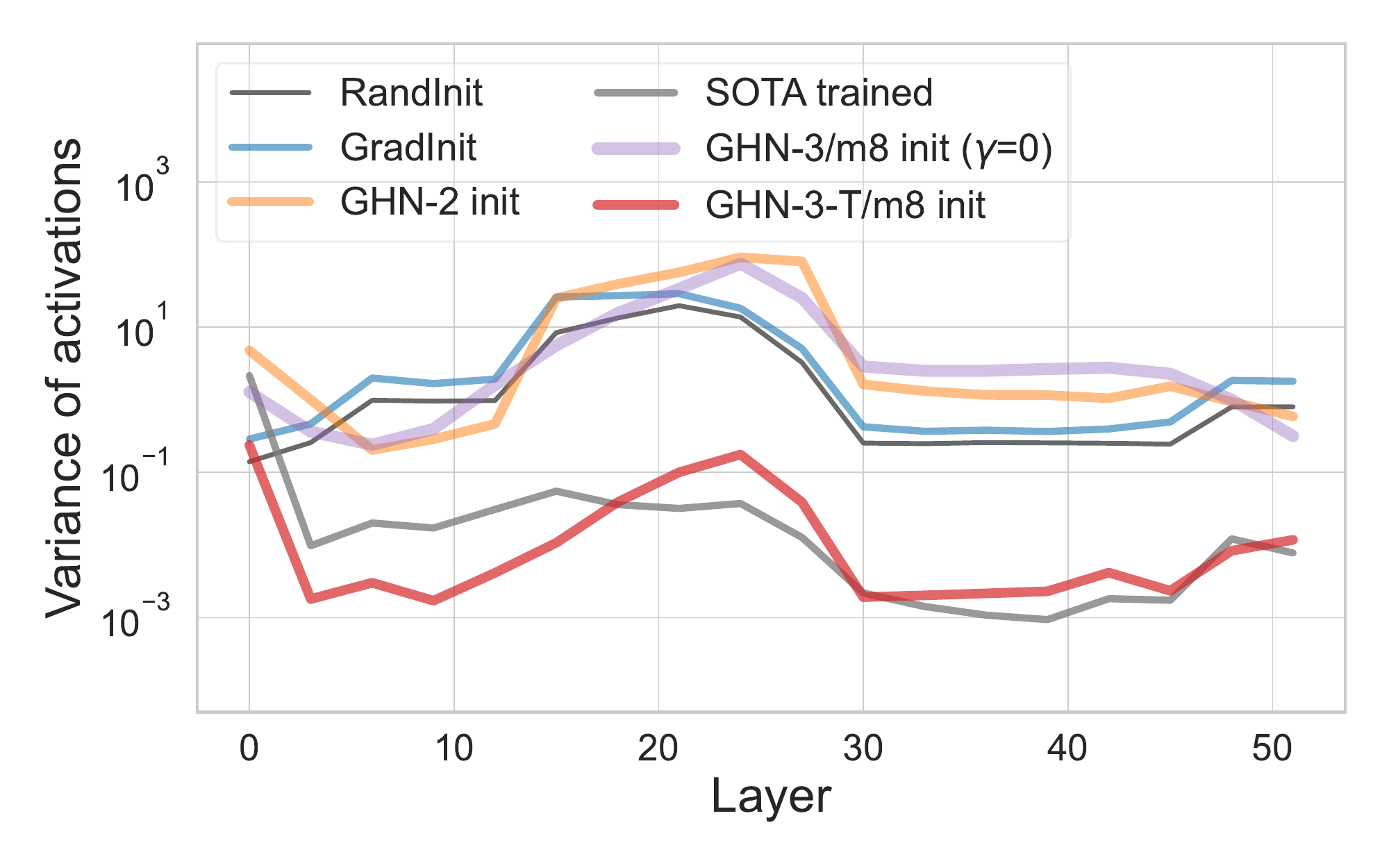} & 
     \includegraphics[width=0.32\textwidth]{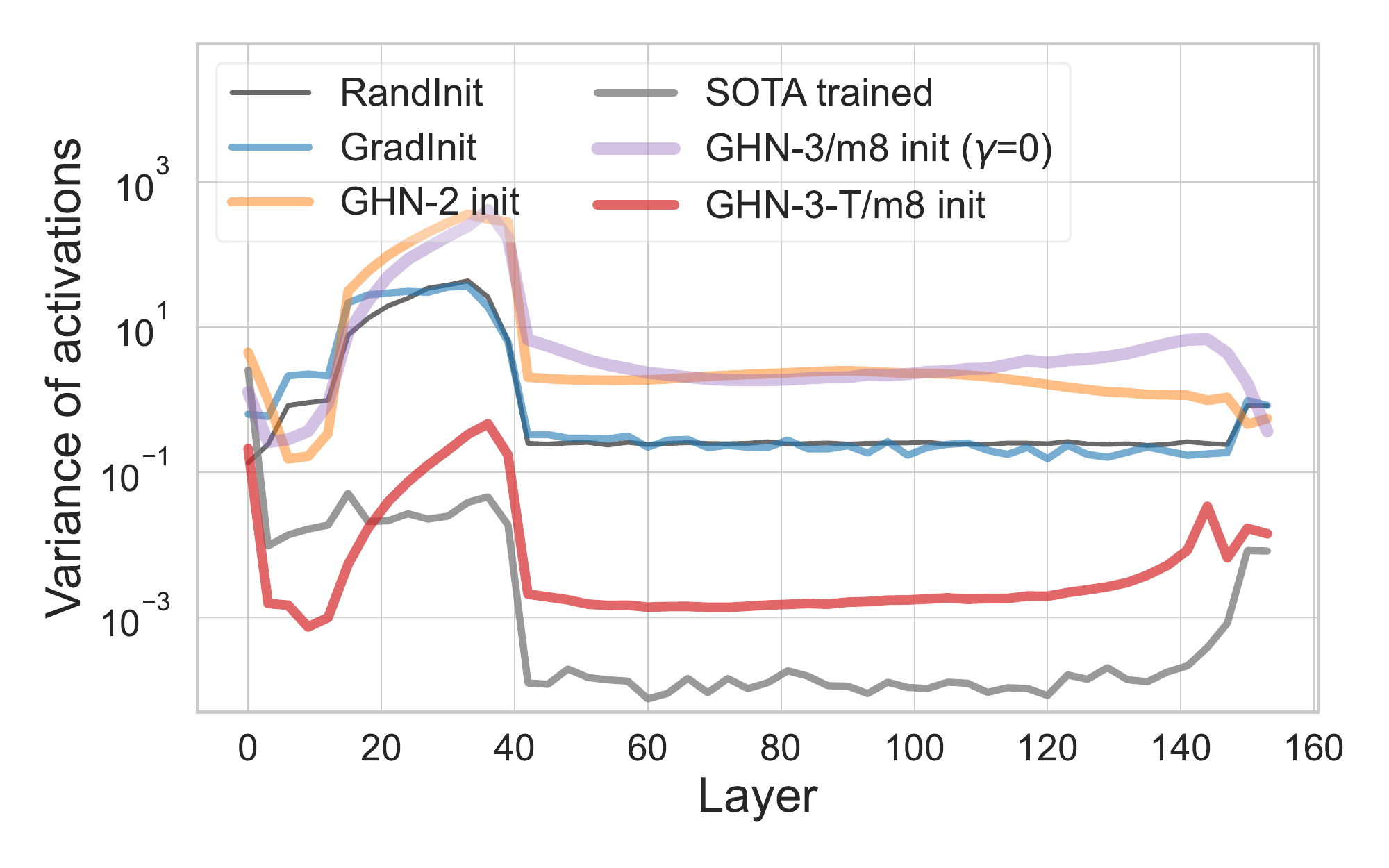} &
     \includegraphics[width=0.32\textwidth]{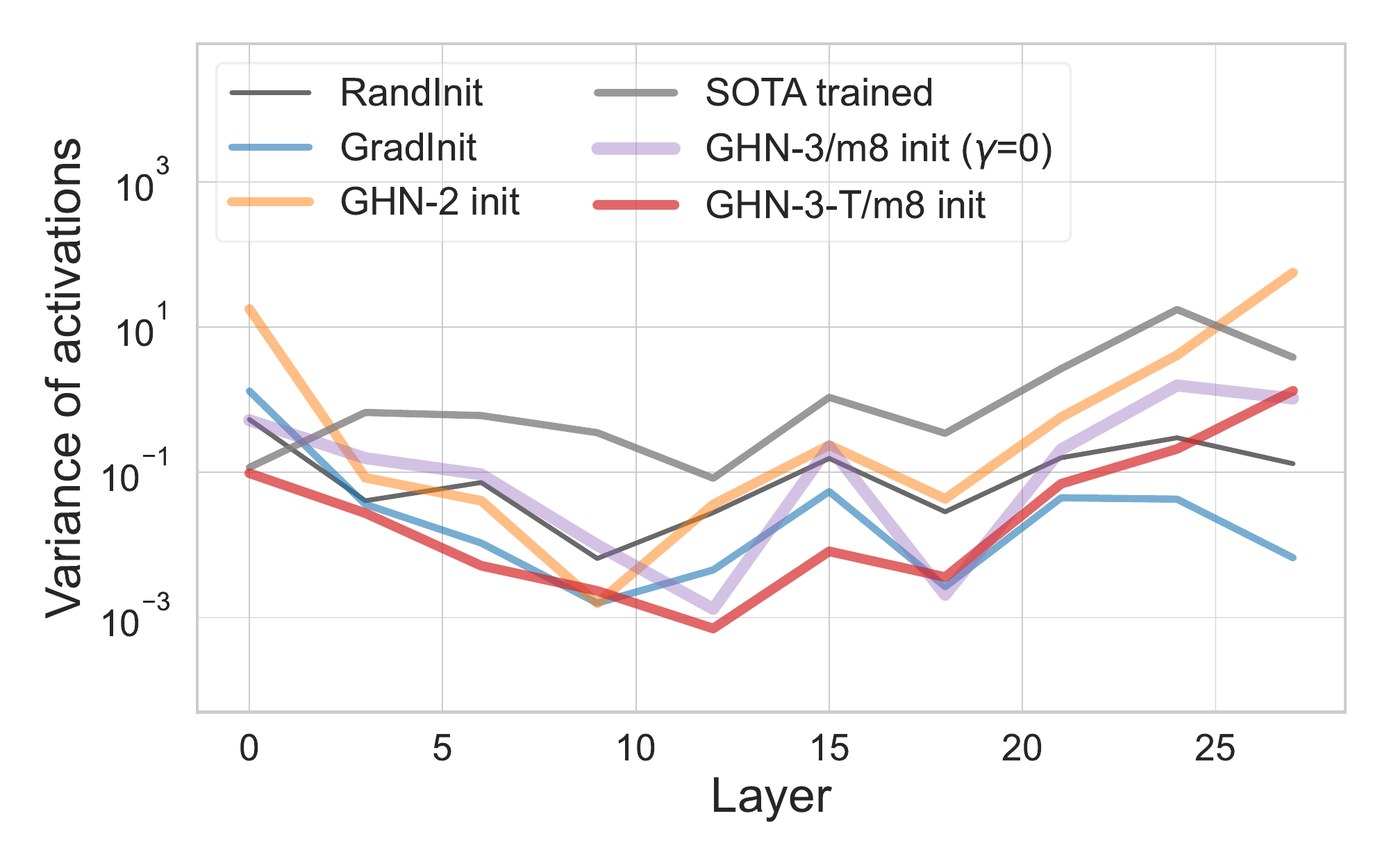}
\end{tabular}
\vspace{-12pt}
\caption{ (\textbf{top}) Validation accuracy curves of ResNet-50, ResNet-152 and Swin-T on ImageNet. 
See the training accuracy curves in Fig.~\ref{fig:train_val_curves} in Appendix. Standard learning rate schedules are used for ResNets (decay every 30 epochs) and Swin-T (cosine decay). (\textbf{bottom}) Variance of activations in the networks initialized with the methods we compare (see Section~\ref{sec:qualitative} for the details).}
\label{fig:val_curves}
\end{center}
\vskip -0.1in
\end{figure*}

\textbf{Generalization.}
When predicted parameters are evaluated without fine-tuning, \ghntwo showed good generalization to unseen architectures of \dataset, including the out-of-distribution splits: Wide, Deep, Dense and BN-Free \cite{knyazev2021parameter}. However, the performance of \ghntwo is still lower compared to \randnospace+SGD-1ep (Fig.~\ref{fig:generalization}).
Our \ghnxl/m16 without any fine-tuning not only matches but outperforms training with SGD for 1 epoch on all evaluations splits of \dataset by a large margin.\looseness-1

\subsection{Full Training}
\label{sec:full_train}

\textbf{Setup.}
We explore if the benefits of a \ghnthree-based initialization still hold when the networks are trained for many epochs as in SOTA training. We compare the \ghnxl/m16-based initialization to \ghntwo, \rand and stronger initialization baselines, \grad~\cite{zhu2021gradinit} and NIO~\cite{yang2022towards}.
We consider three architectures, ResNet-50, ResNet-152~\cite{he2016deep} and Swin Transformer~\cite{liu2021swin} (its tiny variant denoted as Swin-T), achieving strong performance in vision tasks.
We use standard training settings for ResNets:
90 epochs of SGD with momentum 0.9 and batch size 128. A standard (well-tuned) initial learning rate 0.1 is used for \rand, \grad and NIO, while for GHN-based initializations we perform light tuning of the initial learning rate among (0.1, 0.025, 0.01), since predicted parameters may need only slight fine-tuning. In all cases, the learning rate is decayed every 30 epochs.
For Swin-T we follow \citet{liu2021swin} and train with AdamW and the cosine learning rate scheduling, but use computationally light settings to make training feasible. In particular, we use batch size 512 and train for 90 epochs and only using standard augmentation methods as for the ResNets. For Swin-T each initialization method is tuned: initial learning rate is chosen from (1e-3, 6e-4, 3e-4) and the weight decay from (0.01, 0.05).
For all the networks when using the GHN initialization, to break the symmetry of identical parameters, we add a small amount of noise with $\beta$=1e-5 to all parameters following~\citet{knyazev2022pretraining}. We repeat training runs for 3 random seeds in all cases.\looseness-1

\begin{table}[t]
    \centering
    \vspace{-12pt}
    \caption{Comparison of initialization methods for ResNet-50, ResNet-152 and Swin-T on ImageNet. Average accuracies over 3 runs are reported. For all methods, a standard deviation after 1 epoch is $\leq2$, after 45 epochs is $\leq0.3$ and after 90 epochs is $\leq0.1$. Initialization time is measured on 1xNVIDIA-A100 in seconds; average time over the three networks is reported. We also compare to FixUp~\cite{zhang2019fixup} in Table~\ref{tab:no_bn}.\looseness-1
    }
    \label{tab:init_resnet_swint}
    \vskip 0.01in
    \begin{center}
    \begin{scriptsize}
    \begin{sc}
    \setlength{\tabcolsep}{1.7pt}
    \begin{tabular}{lccccp{0.02cm}cccp{0.02cm}ccc}
        \toprule
        \textbf{Init.} & \textbf{Init.} &
        \multicolumn{3}{c}{\textbf{ResNet-50}} & & \multicolumn{3}{c}{\textbf{ResNet-152}} & &
        \multicolumn{3}{c}{\textbf{Swin-T}}\\
        \textbf{method} & \textbf{time} & 1 ep & 45 ep & 90 ep & & 1 ep & 45 ep & 90 ep & & 1 ep & 45 ep & 90 ep \\
        \midrule
        
        \rand & 0.2 & 20.6 & 67.3 & 76.0 & &
        19.7 & 70.5 & \textbf{78.2} & &
        18.4 & 72.7 & 75.6 \\
        
        \grad & 1700 & 21.1 & 67.9 & \textbf{76.2} & & 22.5 & 70.5 & \textbf{78.2} &  &  20.3 & 72.1 & 75.0 \\ 
        
        NIO & 85 & 21.0 & 68.1 & 76.1 & & $-$ & $-$ & $-$ & & $-$ & $-$ & $-$ \\
        
        \ghntwo & 1.3 & 12.8 & 66.5 & 75.6 & & 13.9 & 69.2 & 77.4 & &
        15.8 & 70.3 & 75.8 \\
        
        \ghnthree & 1.2 & \textbf{35.0} & \textbf{73.5} & \textbf{76.2} & & \textbf{34.3} & \textbf{75.9} & 78.1 & &  \textbf{27.1} & \textbf{73.7} & \textbf{76.4}\\
        
         \bottomrule
    \end{tabular}
    \end{sc}
    \end{scriptsize}
    \end{center}
    \vskip -0.2in
\end{table}

\textbf{Results.}
As shown in Fig.~\ref{fig:val_curves}-top and Table~\ref{tab:init_resnet_swint}, our \ghnthree-based initialization speeds up convergence for all the three networks compared to all other initialization methods, including \grad and NIO. 
The gap between our \ghnthree and others is especially high during the first few epochs and particularly for ResNets.
After training for longer, the gap between the methods gradually shrinks and all the initialization methods show comparable performance with two exceptions.
First, on Swin-T the final accuracy of our \ghnthree-based initialization is noticeably better than all other methods (76.4\% vs 75.8\% of the best baseline).
Second, the baseline \ghntwo is generally worse than other initialization methods converging slowly and underperforming at the end.
Overall, our \ghnthree brings GHNs to a significantly better level by improving convergence versus other methods in the first epochs.
Advanced initialization methods, NIO and \grad, performed similarly to each other in our ResNet-50 experiments, and we were unable to run NIO for the other networks (Table~\ref{tab:init_resnet_swint}).
Compared to NIO and \grad, \ghnthree has an extra advantage besides converging faster in the beginning. Specifically, NIO and \grad require propagating and computing gradients for 100-2000 mini-batches on ImageNet to initialize each network, which can be inconvenient in practice~\cite{zhu2021gradinit,yang2022towards}. In contrast, our trained \ghnthree predicts ImageNet parameters for each network in $\approx$ 1 second (even on a CPU) and without accessing ImageNet.\looseness-1

\textbf{Modern training recipes.}
While we used standard training recipes with 90 epochs, modern training recipes require up to 600 epochs with stronger regularizations and other tricks to reach higher accuracy at a higher computational cost~\cite{wightman2021resnet}.
However, the goal of this work is \textit{not} to achieve SOTA at a high cost, but to achieve competitive performance at a low computational cost (in a few epochs) by leveraging \ghnthree. As we report in Table~\ref{tab:init_resnet_swint}, the final 90 epoch accuracies of the \ghnthree-based initialization are generally similar to \rand and we expect this trend to maintain if the modern recipes are used.

\subsection{Transfer Learning}

\textbf{Setup.}
We explore if the parameters predicted by our GHNs for ImageNet can be transferred to other tasks.
We consider three architectures: ResNet-50, Swin-T and the base variant of ConvNeXt~\cite{liu2022convnet}.
We compare the following initialization approaches: (1) \rand, (2) orthogonal initialization~\cite{saxe2013exact}, (3) parameters predicted by GHNs, (4) \rand or predicted parameters trained/fine-tuned on ImageNet for 1 epoch and (5) \rand trained on ImageNet for 90-600 epochs. The latter requires significant computational resources and therefore is not considered to be a fair baseline. We fine-tune the networks initialized with one of these approaches on the few-shot variants of the CIFAR-10 and CIFAR-100 image classification tasks~\cite{krizhevsky2009learning}. Following~\cite{zhai2019large,knyazev2021parameter} we consider 1000 training samples in each task, which is well suited to study transfer learning abilities.
To transfer ImageNet parameters to CIFAR-10 and CIFAR-100, we re-initialize the classification layer with \rand with 10 and 100 outputs respectively and fine-tune the entire network.
We tune hyperparameters for each method in a fair fashion following~\cite{knyazev2022pretraining}. We also evaluate on the Penn-Fudan object detection task~\cite{wang2007object} containing only 170 images. We closely follow the setup and hyperparameters from \cite{knyazev2021parameter,pytorchdetection} and evaluate on three common backbones, ResNet-50, ResNet-101 and ResNet-152~\cite{he2016deep}.
In all cases, we repeat experiments 3 times and report an average and standard deviation.\looseness-1

\begin{table}[t]
    \vspace{-5pt}
    \centering
    \caption{Transfer learning from ImageNet to few-shot CIFAR-10 and CIFAR-100 with 1000 training labels with 3 networks: ResNet-50 (R-50), ConvNext-B (C-B) and Swin-T (S-T).  Average accuracies over 3 runs are reported; std in all cases is generally $\leq$ 0.5.\looseness-1
    }
    \label{tab:ft}
    \vspace{-2pt}
    \begin{center}
    \begin{scriptsize}
    \begin{sc}
    \setlength{\tabcolsep}{3.5pt}
    \begin{tabular}{lc|ccc|ccc} 
        \toprule
        \textbf{Initialization} &
        \textbf{\tiny ImageNet} &
        \multicolumn{3}{c|}{\textbf{CIFAR-10}} & \multicolumn{3}{c}{\textbf{CIFAR-100}} \\
        
       & \textbf{\tiny pretrain} & R-50 & C-B & S-T & R-50 & C-B & S-T \\
        \midrule
        
        \rand & No & 61.6 & 48.0 & 46.0 & 14.5 & 11.6 & {12.2} \\
        
        Orth & No & 61.1 & 52.4 & 47.8 & 14.8 & 13.6 & 12.5 \\
        
        \ghntwo & No & 61.8 & 52.3 & 48.2 & 18.3 & {13.7} & \textbf{12.7} \\
        
        \ghnthree-XL/m16 & No & \textbf{72.9} & \textbf{55.3} & \textbf{51.8} & \textbf{27.8} & \textbf{13.8} & {11.9}  \\
        
        \midrule
        \rand & 1 epoch & 74.0 & 69.1 & 62.1 & 33.1 & 19.9 & 24.2 \\
        
        \ghntwo & 1 epoch & 68.4 & 69.1 & 58.6 & 26.6 & 28.0 & 21.4\\
        
        \ghnthree-XL/m16 & 1 epoch & \textbf{77.8} & \textbf{71.3} & \textbf{64.3} & \textbf{37.2} & \textbf{31.0} & \textbf{26.5}\\

        \midrule
        \midrule
        \textcolor{gray}{\rand} & \textcolor{gray}{90-600 ep} &  \textcolor{gray}{88.7} & \textcolor{gray}{95.6} & \textcolor{gray}{93.4} & \textcolor{gray}{56.1} & \textcolor{gray}{69.7} & \textcolor{gray}{62.5}  \\
         \bottomrule
    \end{tabular}
    \end{sc}
    \end{scriptsize}
    \end{center}
    \vskip -0.2in
\end{table}

\begin{table}[t]
    \vspace{-5pt}
    \centering
    \caption{Transfer learning results (average precision at IoU=0.50 measured in \%) for the Penn-Fudan object detection dataset. 
    }
    \label{tab:ft_object}
    \vspace{-2pt}
    \begin{center}
    \begin{scriptsize}
    \begin{sc}
    \setlength{\tabcolsep}{2pt}
    \begin{tabular}{lcccc} 
        \toprule
        \textbf{Initialization} &
       \textbf{\tiny ImNet pretr.} &
        \textbf{ResNet-50}  & \textbf{ResNet-101} & \textbf{ResNet-152} \\
        \midrule
        
        \rand & No & 21.3\std{4.9} & 14.9\std{1.1} & 18.4\std{1.9}\\
        
        \ghntwo & No & 54.9\std{1.4} & 55.1\std{3.2} & 55.8\std{5.7} \\
        
        \ghnthree-XL/m16 & No & \textbf{61.7\std{3.8}} & \textbf{60.2\std{7.6}} & \textbf{60.0\std{5.2}} \\
        
        \midrule
        \midrule
        \textcolor{gray}{\rand} & \textcolor{gray}{90 epochs} &  \textcolor{gray}{87.6\std{1.1}} & \textcolor{gray}{88.2\std{4.8}} & \textcolor{gray}{89.3\std{5.1}}\\
         \bottomrule
    \end{tabular}
    \end{sc}
    \end{scriptsize}
    \end{center}
    \vskip -0.2in
\end{table}

\textbf{Results.} The parameters predicted by our GHNs show better transferability compared to GHN-2 in all but one of the transfer learning experiments (Tables~\ref{tab:ft} and \ref{tab:ft_object}). We also significantly improve on \rand. By fine-tuning the predicted parameters on ImageNet for 1 epoch before transferring them to the CIFAR datasets we achieve further boosts and outperform \randnospace+SGD-1ep. While the gap with full ImageNet pretraining is still noticeable (e.g. 77.8\% vs 88.7\% for ResNet-50 on CIFAR-10), we narrow this gap compared to \ghntwo (68.4\%) and \rand  (74.0\%).

\subsection{Qualitative Analysis and Discussion}
\label{sec:qualitative}

We analyze the diversity of predicted parameters following experiments in~\cite{knyazev2021parameter}.
We predict parameters for all \textsc{PyTorch} models and from them collect all the parameter tensors of one of the three frequently occurring shapes.
We then compute the absolute cosine distance between all pairs of parameter tensors of the same shape and report the mean of the pairwise matrix (Table~\ref{tab:diversity}). 
In general the parameters predicted by our GHN-3 are more diverse than the ones predicted by \ghntwo even though we did not enforce the diversity during training \ghnthree. One exception is parameters of the shape $64\PLH3\PLH7\PLH7$ used in the first convolutional layer of many models. However, in both \ghntwo and \ghnthree the diversity of this tensor is small, while \ghnthree improves the quality visually  (Fig.~\ref{fig:resnet50-conv1}).\looseness-1

We also evaluate the quality of predicted parameters by computing the variance of activations when propagating a mini-batch of images through the network. We initialize the networks with random-based methods, SOTA trained parameters or parameters predicted by GHNs and propagate the same mini-batch of images through the network to compute activations and their variances (Fig.~\ref{fig:val_curves}-bottom). For the ResNets, the \ghnthree initialization leads to the variances well aligned with SOTA trained models. Our predicted parameter regularization (Section~\ref{sec:wd}) is important to enable this behavior.
However, our activations are not always well aligned with pretrained models as we show for ViT and EfficientNet (Fig.~\ref{fig:variances_vit} in Appendix), which may explain a lower fine-tuning accuracy for these networks (Fig.~\ref{fig:pytorch}-right).

Finally, GHNs can be used to efficiently estimate the performance of architectures making them a potentially useful approach for neural architecture search (NAS). However, GHNs are not explicitly trained to perform the NAS task and in our experiments underperformed compared to other NAS methods (Appendix~\ref{apdx:nas}). At the same time, our \ghnxl/m16 outperformed \ghntwo and a smaller scale \ghnthree indicating the promise of large-scale GHNs for NAS.

We provide additional results and discussion in Appendix~\ref{sec:apdx}.

\begin{table}[t]
    \vspace{-5pt}
    \centering
    \caption{Diversity of the parameters predicted by GHNs \textit{vs} SOTA trained by SGD from \rand measured on the \textsc{PyTorch} split. \textsuperscript{*}Since the parameters in the SOTA trained models are not ordered in a canonical way, we employ the Hungarian matching~\cite{kuhn1955hungarian} before computing the distance between a pair of tensors.
    }
    \label{tab:diversity}
    \vspace{-2pt}
    \begin{center}
    \begin{scriptsize}
    \begin{sc}
    \setlength{\tabcolsep}{4pt}
    \begin{tabular}{lccc}
        \toprule
        \textbf{Method} & \multicolumn{3}{c}{\textbf{Parameter tensor shape}} \\
        & \tiny $64\PLH3\PLH7\PLH7$ & \tiny $256\PLH256\PLH3\PLH3$ & \tiny $1024\PLH1024\PLH1\PLH1$ \\
        \midrule
        
        MLP & 0.0 & 0.174 & 0.020\\
        
        GHN-2 &  \textbf{1e-3} & 0.283 & 0.070 \\
        \ghnxl/m16 & 2e-4 & \textbf{0.310} & \textbf{0.095} \\
        
        \midrule

        \textcolor{gray}{SOTA training\textsuperscript{*}} & 
        \textcolor{gray}{0.388} & \textcolor{gray}{0.917} & \textcolor{gray}{0.895} \\

         \bottomrule
    \end{tabular}
    \end{sc}
    \end{scriptsize}
    \end{center}
    \vskip -0.25in
\end{table}

\section{Conclusion}
\vspace{-3pt}
We improve Graph HyperNetworks by considerably scaling them up. By evaluating on realistic and challenging ImageNet architectures, we found that scaling up gradually increases the overall performance. This is encouraging as further scaling GHNs can make them a powerful tool. Our \ghnthree improves random-based and advanced initializations in vision experiments and makes a big step in the quality of predicted parameters compared to the prior GHNs.

\section*{Acknowledgements}
\vspace{-7pt}
The experiments were in part enabled by computational resources provided by Calcul Quebec and Compute Canada. 
Simon Lacoste-Julien is a CIFAR Associate Fellow in the Learning in Machines \& Brains program.

\setlength{\bibsep}{6.5pt plus 1ex}
\bibliography{ref}
\bibliographystyle{icml2023}

\newpage
\appendix



\section{APPENDIX}
\label{sec:apdx}

\subsection{GHN Details}
\label{apdx:details}

The decoder of our \ghnthree has the same architecture as in \ghntwo, however when we scale GHN-3 we set the final output dimensionality of the decoder to $d \PLH d \PLH 16 \PLH 16$ instead of $2d \PLH 2d \PLH 16 \PLH 16$.
This final output dimensionality determines the maximum size of the predicted parameter tensor and larger/smaller tensors are obtained by tiling/slicing the largest one following the implementation in~\cite{knyazev2021parameter}.
See Section~\ref{apdx:results} for an empirical analysis of this design choice.

In addition to the in-degree and out-degree (centrality) node embeddings (Eq.~\ref{eq:embed_base}) our GHN-3 models add one extra embedding to nodes. In particular, to each node we add the embedding corresponding to the distance from the input node $i=0$. This embedding is not important for our results, and we empirically analyze the usefulness of the centrality and input distance embeddings in Section~\ref{apdx:results}.

For a few recent architectures (e.g. ConvNeXt) in the \textsc{PyTorch} split, some layers are not supported by GHNs (not available during training). We do not predict the parameters of those layers and use standard initialization for them.

\subsection{Predicted Parameter Regularization Ablations}
\label{apdx:ablations}

Alternatively to applying our predicted parameter regularization (Eq.~\ref{eq:wd}), it is possible to apply a higher weight decay $\lambda$ on the GHN parameters to implicitly enforce it predict smaller values. We verified that increasing $\lambda$ by a factor of 10 only slightly improved the results of the GHN with $\gamma$=0 (from 19.9\% to 21.3\%, see Table~\ref{tab:ablations_more}), while further increasing $\lambda$ to 0.3 worsens results  (from 19.9\% to 18.9\%). The results with increased $\lambda$ confirm the advantage of explicitly regularizing predicted parameters (23.9\%). Tuning $\gamma$ is also important and using a larger value (1e-4 instead of 3e-5) worsens results (20.0\% vs 23.9\%).

\begin{table}[th]
    \centering
    \caption{Additional ablation results following Table~\ref{tab:ablations}.}
    \label{tab:ablations_more}
    \vskip 0.05in
    \begin{center}
    \begin{scriptsize}
    \begin{sc}
    \setlength{\tabcolsep}{1pt}
    \begin{tabular}{lcc}
        \toprule
        \textbf{Model} & 
        \textbf{Params}
        & \textbf{Acc}\\
        \midrule
        \ghnt/m8 ($\gamma$=3e-5, $\lambda$=0.01) & 6.91M & 23.9\std{1.0} \\
        \midrule
        
        No predicted param. reg. ($\gamma$=0) in Eq.~\eqref{eq:wd} & 6.91M & 19.9\std{4.7} \\
        
        No predicted param. reg. ($\gamma$=0) in Eq.~\eqref{eq:wd}, $\lambda$=0.1 & 6.91M & 21.3\std{1.9} \\
        
        No predicted param. reg. ($\gamma$=0) in Eq.~\eqref{eq:wd}, $\lambda$=0.3 & 6.91M & 18.9\std{1.1} \\
        
        Predicted param. reg. $\gamma$=1e-4 in Eq.~\eqref{eq:wd} & 6.91M & 20.0\std{3.0} \\ 
        
        \bottomrule
    \end{tabular}
    \end{sc}
    \end{scriptsize}
    \end{center}
\end{table}

\subsection{Additional Results}
\label{apdx:results}

\subsubsection{Additional GHN variants}
In Table~\ref{tab:generalization_more} we report the results for more GHN variants on \dataset without fine-tuning predicted parameters.

First, we retrained \ghntwo using our implementation and hyperparameters (see \textbf{iii} in Table~\ref{tab:generalization_more}), including predicted parameter regularization (Eq.\ref{eq:wd}). It performed slightly better than the \ghntwo released by \citet{knyazev2021parameter} (\textbf{ii}) implying that our hyperparameters (optimizer, learning rate, etc.) are preferable. At the same time, due to our implementation improvements it takes two times less time to train it. We also trained a smaller variant of our \ghnt (\textbf{iv}) with the same hidden size $d$ and same decoder output shape ($2d \PLH 2d \PLH 16 \PLH 16$) as in \ghntwo, which has a comparable number of trainable parameters as the baseline \ghntwo. This \ghnthree variant performs slightly better than the retrained \ghntwo, but takes 3.5 times less time to train than \ghntwo (\textbf{iii}).

We also trained \ghntwo with a larger hidden size ($d$=128, see\textbf{x} in Table~\ref{tab:generalization_more}), which has about the same number of parameters as our \ghns (\textbf{xii}). It performed worse than \ghns and is about three times longer to train.
Overall, \ghnthree is much faster to train than \ghntwo and is more performant due to a deep Graphormer-based architecture rather than a shallow GatedGNN-based architecture. These two factors allow us to scale up \ghnthree and obtain significant improvements.
We also compared \ghnt with our default output dimensionality $d \PLH d \PLH 16 \PLH 16$ (\textbf{xii}) to a GHN with a larger output dimensionality, $4d \PLH 4d \PLH 16 \PLH 16$ (\textbf{xi}). The latter performed worse compared to \ghns with around the same number of parameters validating our approach to scale up GHNs.

Finally, we evaluate a GHN without centrality node embeddings introduced in Graphormers (Eq.~\ref{eq:embed_base}) and found that its usefulness is limited (see \textbf{v} and \textbf{vii} in Table~\ref{tab:generalization_more}).
Regarding the input distance embedding (Section~\ref{apdx:details}), we found it to be useful in the initial experiments. However similarly to the centrality embedding, in our final experiments it provided only marginal gains compared to the \ghnt without the input distance embedding (see \textbf{v} and \textbf{vii}).
At the same time, these node embeddings become very useful when edge embeddings are removed (see \textbf{viii} and \textbf{ix} and \textbf{vii}). Apparently, these node embeddings provide some noisy information about the graph structure, but this information becomes less useful when edge embeddings are added.
Since both the centrality and input distance embeddings do not introduce significantly more trainable parameters or computational demands while provide small gains in some cases, we keep them in our final GHNs.

\subsubsection{Results summary}
\begin{figure}[t]
\begin{center}
\centerline{\includegraphics[width=0.99\columnwidth]{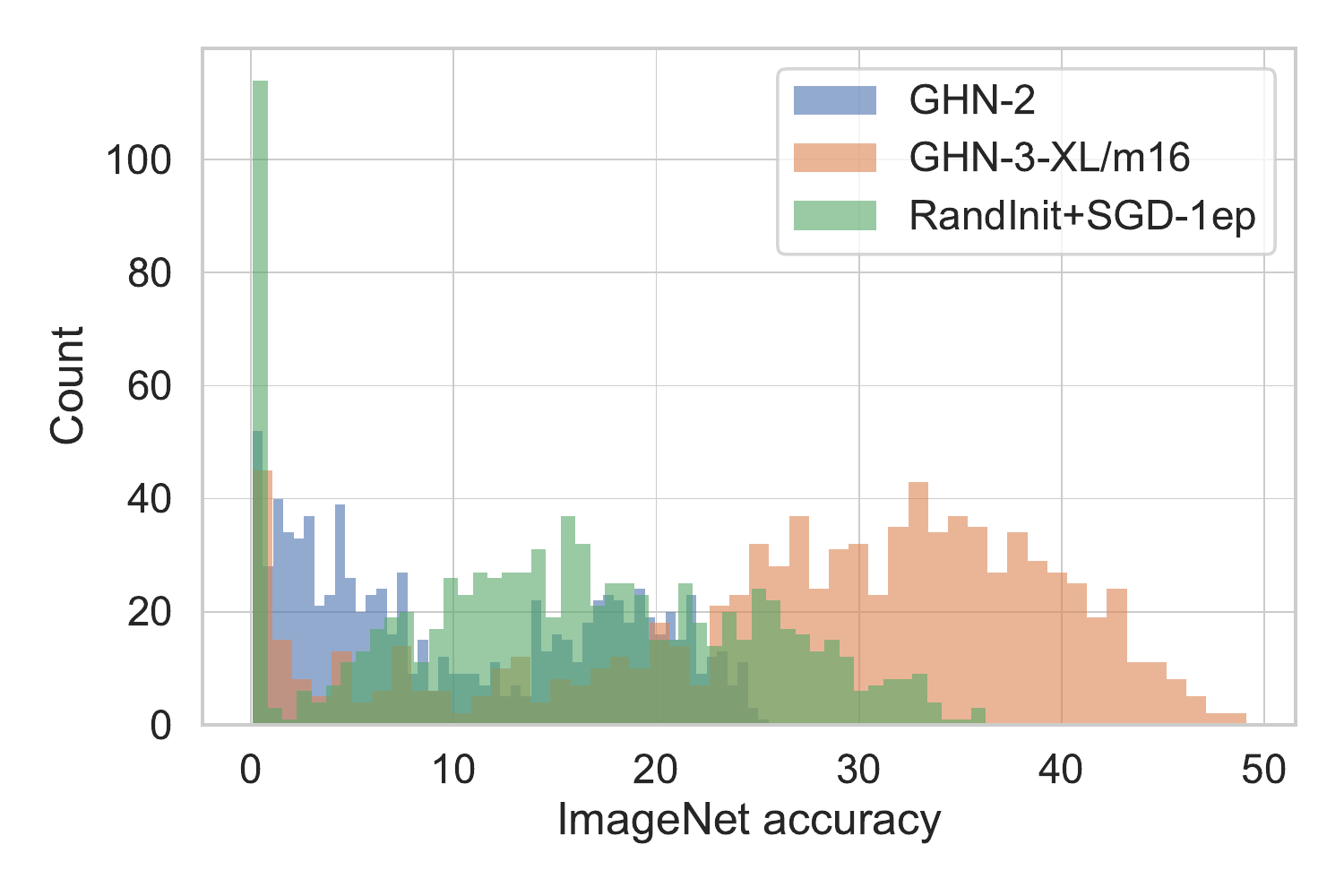}}
\vskip -0.2in
\caption{ImageNet top-1 accuracy on all the 900 architectures of \dataset for GHNs \textbf{without fine-tuning} vs \randnospace+SGD-1ep.
}
\label{fig:deepnets}
\end{center}
\vskip -0.2in
\end{figure}

Table~\ref{tab:summary_all} reports the results of GHNs vs \rand on all 900 + 74 evaluation networks without and with fine-tuning. These results summarize the histograms in Fig.~\ref{fig:pytorch}. The distribution of accuracies on \dataset is shown in Fig.~\ref{fig:deepnets}.

\begin{table*}[h]
    \centering
    \caption{Generalization results on the evaluation architecture splits of \dataset using ImageNet top-1 accuracy. \textsuperscript{*}Days on 4xNVIDIA-A100 for 75 epochs. \textsuperscript{**}Days on 8xNVIDIA-A100 for 75 epochs.
    }
    \label{tab:generalization_more}
    \vspace{1pt}
    \begin{center}
    \begin{tiny}
    \begin{sc}
    \setlength{\tabcolsep}{2pt}
    \rowcolors{4}{white}{gray!10}
    \begin{tabular}{llcccccccc|c}
        \toprule
        \textbf{\#} & \textbf{Method} & \textbf{SGD steps} & \textbf{Params (M)} & \textbf{Train time\textsuperscript{*}} &
        \textbf{\iidtest} & \textbf{\wide} & \textbf{\deep} &
         \textbf{\dense} & \textbf{\bnfree} & \textbf{All} \\
        \midrule
        & \# architectures & & & & 500 & 100 & 100 & 100 & 100 & 900\\
        
        (i) & \rand & 10k (1ep) & $-$ & $-$ & 17.7\std{7.7} & 18.9\std{9.9} & 9.1\std{7.3} & 16.8\std{7.7} & 3.2\std{5.2} & 15.1\std{9.2}\Tstrut\\

        (ii) & \ghntwo/m8 & 0 & 2.32 & 21.9 & 12.1\std{7.6} & 7.9\std{7.2} & 10.6\std{7.1} & 11.3\std{6.7} & 2.2\std{1.9} & 10.3\std{7.6}\\
        
        (iii) & \ghntwo/m8 (our implem. and hyperparams and Eq.\eqref{eq:wd}) & 0 & 2.32 & 11.6 & 14.3\std{6.5} & 15.2\std{6.4} & 12.7\std{6.5} & 12.2\std{6.5} & 2.9\std{3.0} & 12.7\std{7.2}\\
        
        (iv) & \ghnthree/m8 ($d=32$, decoder output: $2d \PLH 2d \PLH 16 \PLH 16$) & 0 & 2.37 & 3.2 & 14.3\std{7.4} & 15.5\std{7.3} & 13.0\std{6.8} & 12.8\std{6.7} & 2.1\std{2.4} & 12.8\std{7.9} \\
        
        \midrule
        
        (v) & \ghnt/m8-no centr. embed. (Eq.~\ref{eq:embed_base}) & 0 & 6.89 & 3.8 & 18.6\std{7.7} & 20.2\std{7.4} & 17.8\std{7.7} & 16.9\std{8.1} & 6.4\std{3.6} & 17.1\std{8.3}\Tstrut\\
        
        (vi) & \ghnt/m8-no input dist. embed. (Section~\ref{apdx:details}) & 0 & 6.84 & 3.8 & 19.0\std{7.7} & 20.1\std{8.2} & 18.0\std{7.9} & 17.0\std{8.4} & 6.5\std{3.8} & 17.4\std{8.5} \\
        
        (vii) & \ghnt/m8 & 0 & 6.91 & 3.8 & 19.0\std{7.6} & 20.5\std{7.5} & 18.0\std{7.7} & 16.9\std{8.3} & 6.9\std{4.1} & 17.5\std{8.3}\\
        
        \midrule
        
        (viii) & \ghnt/m8, No fw/bw embed. in Eq.~\eqref{eq:sa} & 0 & 6.88 & 3.7 & 14.7\std{6.5} & 15.3\std{7.0} & 13.5\std{6.2} & 13.0\std{6.6} & 4.4\std{3.8} & 13.3\std{7.1}\Tstrut\\
        
        (ix) & \ghnt/m8, No fw/bw/centr/input dist. embed & 0 & 6.80 & 3.7 & 10.2\std{5.5} & 9.5\std{6.0} & 9.5\std{4.8} & 9.3\std{4.8} & 4.0\std{3.0} & 9.3\std{5.5} \\
        \midrule
        
        (x) & \ghntwo/m8 ($d$=128) (our implem. and hyperparams) & 0 & 34.73 & 11.9 & 22.4\std{9.8} & 22.2\std{11.3} & 21.7\std{9.3} & 18.5\std{9.9} & 6.8\std{6.0} & 20.1\std{10.8} \\

        (xi) & \ghnt/m8 (decoder output: $4d \PLH 4d \PLH 16 \PLH 16$) & 0 & 38.86 & 3.8 & 23.9\std{10.8} & 21.5\std{13.7} & 22.9\std{9.6} & 20.2\std{10.5} & 7.6\std{6.2} & 21.3\std{11.7}\\

        (xii) & \ghns/m8 & 0 & 35.81 & 3.8 & \textbf{24.2\std{10.0}} & \textbf{25.7\std{11.3}} & \textbf{23.8\std{9.5}} & \textbf{21.2\std{10.1}} & \textbf{10.3\std{5.2}} & \textbf{22.4\std{10.7}}\\
        
        \midrule
        
        (xiii) & \ghnxl/m16 & 0 & 654.37 & 7.1\textsuperscript{**} & \textbf{29.8\std{11.2}} & \textbf{26.8\std{15.6}} & \textbf{28.8\std{9.5}} & \textbf{26.2\std{10.9}} & \textbf{15.1\std{9.7}} & \textbf{27.3\std{12.3}} \\
        
         \bottomrule
    \end{tabular}
    \end{sc}
    \end{tiny}
    \end{center}
    \vskip -0.1in
\end{table*}

\begin{table}[t]
    \centering
    \caption{ImageNet top-1 accuracy for all 900 + 74 evaluations networks in \dataset and \textsc{PyTorch}.
    Average and standard deviation of accuracies for all networks in each split are reported.
    }
    \label{tab:summary_all}
    \vskip 0.15in
    \begin{center}
    \begin{scriptsize}
    \begin{sc}
    \begin{tabular}{lccc}
        \toprule
        & \textbf{SGD steps} & \textbf{\dataset} & \textbf{PyTorch} \\
        \midrule
        \ghntwo & 0 & 10.3\std{7.6} & 0.2\std{0.3}\\
        \ghnxl/m16 & 0 & 27.3\std{12.3} & 1.7\std{5.0} \\
        \midrule
        \rand & 10k (1ep) & 15.1\std{9.2} & 17.4\std{7.2} \\
        
        \ghnxl/m16 & 1k (1/10ep) & $-$ & 8.3\std{10.2}  \\
        \ghnxl/m16 & 10k (1ep) & $-$ & 25.4\std{12.3} \\
         \bottomrule
    \end{tabular}
    \end{sc}
    \end{scriptsize}
    \end{center}
    \vskip -0.1in
\end{table}

\subsubsection{Huge ResNet and ViT}
The main value of our \ghnthree is in strong initialization for many networks that do not have ImageNet weights. For example, for all the 900 evaluation architectures in DeepNets-1M there are no pretrained ImageNet weights and our GHN-3 achieves strong results for them (see Table~\ref{tab:summary_all}).
To further support our argument, we show that we can predict good parameters for huge ResNet-1000 with $>$ 400M parameters and ViT-1.2B with $>$ 1.2 billion parameters (the ViT is based on the ``no distribution shift'' experiment described in Section~\ref{apdx:distr_shift}). No pretrained weights are available for them. We train them for 1 epoch (due to high cost) with \rand and our GHN-3 init. using two learning rates.
We found that GHN-3 can initialize well such huge networks despite being trained on much smaller ones (Table~\ref{tab:huge_resnet_vit}). On average, our GHN-3 init is better for both the 0.1 and 0.01 learning rates and both architectures, in some cases by a large margin, further confirming the strength of our GHN in such a challenging setting.
Potentially, this can lead to significant cost reductions.
GHN-3 also provides more stable gradient norm at initialization (very large gradient norm of \rand for ResNet-1000 may lead to numerical overflow and training instabilities).
Finally, the GHN-3-based initialization can be less sensitive to the learning rate (also see Table~\ref{tab:lr_sensitivity} for additional learning rate sensitivity results).

\begin{table}[thpb]
    \centering
    \caption{ImageNet top-1 accuracy for huge ResNet and ViT after 1 epoch of training (averaged for 3 runs). }
    \label{tab:huge_resnet_vit}
    \begin{center}
    \begin{scriptsize}
    \begin{sc}
    \setlength{\tabcolsep}{2pt}
    \begin{tabular}{llccc}
    \toprule
     \textbf{Model} & \textbf{Init.} & \textbf{initial grad norm} & \multicolumn{2}{c}{\textbf{1 epoch}} \\
     & & & lr=0.1 & lr=0.01\\
     \midrule
    \multirow{2}{*}{ResNet-1000} & \rand & 99121.9 & 21.0$\pm$1.0 & 15.5$\pm$0.7 \\
     & \ghnthree & 72.0 & \textbf{23.1$\pm$1.9} & \textbf{32.2$\pm$1.1} \\
    \midrule
    \multirow{2}{*}{ViT-1.2B} & \rand & 4.8 & 2.0$\pm$0.5 & 16.5$\pm$0.3 \\
     & \ghnthree & 6.9 & \textbf{11.8$\pm$0.8} & \textbf{17.4$\pm$0.6} \\
    \bottomrule
    \end{tabular}
    \end{sc}
    \end{scriptsize}
    \end{center}
    \vskip -0.1in
\end{table}

\subsubsection{Architecture distribution shift}
\label{apdx:distr_shift}

For some PyTorch architectures, in particular ViT and EfficientNet, there is a certain distribution shift from the DeepNets-1M training architectures used to train GHNs and their specification in PyTorch.

We focus on ViT and EfficientNet as concrete examples:
\begin{itemize}
    \item \textbf{ViT:} ViTs in PyTorch use a classification token, which is never used in the DeepNets-1M architectures. In DeepNets-1M, global average pooling (GAP) is used instead to obtain the last layer features.
    \item \textbf{EfficientNet:} EfficientNets in PyTorch have a squeeze and excitation operation with a SiLU activation, while in DeepNets-1M only ReLU is used in those layers.
\end{itemize}

These distribution shifts can be fixed by either including a PyTorch based classification token and SiLU in the DeepNets-1M training set or by slightly adjusting ViT/EfficientNet architectures in PyTorch. To support our argument, we follow the latter approach. For ViT we replace the classification token with GAP, while for EfficientNet we replace SiLU with ReLU. We report the results after 1 epoch with distr. shift (original PyTorch architectures) and no distr. shift (our adjusted architectures) in Table~\ref{tab:init_vit_efficientnet}.
We show that when the distribution shift is removed, the GHN-based initialization outperforms \rand. We note that the distribution shift issue is inherent to neural network based methods and it is likely to come up with examples where a neural net fails. We did not alter the original PyTorch ViT/EfficientNet architectures in our main experiments in Table~\ref{tab:bench_imagenet} to show this limitation of GHNs. But here we show that the distribution shift problem can be mitigated if needed. We believe future work can focus on designing more diverse training architectures than DeepNets-1M, so that the issue of the distribution shift is less pronounced.

\begin{table}[thpb]
    \centering
    \caption{ImageNet top-1 accuracy for ViT/EfficientNet with and without the distribution shift.}
    \label{tab:init_vit_efficientnet}
    \begin{center}
    \begin{scriptsize}
    \begin{sc}
    \setlength{\tabcolsep}{2pt}
    \begin{tabular}{llcc}
    \toprule
     \textbf{Model} & \textbf{Init. method} & \textbf{With distr. shift} & \textbf{No distr. shift} \\
     \midrule
    \multirow{2}{*}{ViT-B/16} & \rand & \textbf{11.06} & 9.87 \\
     & \ghnthree & 8.49 & \textbf{11.47} \\
    \midrule
    \multirow{2}{*}{EfficientNet-B0} & \rand & \textbf{24.64} & 23.29 \\
     & \ghnthree & 22.19 & \textbf{32.00} \\
    \bottomrule
    \end{tabular}
    \end{sc}
    \end{scriptsize}
    \end{center}
    \vskip -0.1in
\end{table}

\subsubsection{Ensembling}

We evaluated if the predicted parameters of different architectures can be ensembled to improve performance.
We chose three networks for which \ghnthree predicts the parameters that achieve good accuracy without any training (Table~\ref{tab:ensembling}). We construct an ensemble by averaging the logits from the three networks when we propagate the validation images of ImageNet. As another experiment, we also fine-tuned the three networks with SGD for 1 epoch and ensembled them to see how the results change after fine-tuning. We compare to the baseline \ghntwo, MLP, \ghnt (our tiny \ghnthree) and \rand. We report the validation accuracy of individual networks and their ensemble (Table~\ref{tab:ensembling}).
We found that our best model (\ghnthree) predicts somewhat diverse parameters as the results of the ensemble slightly improve even without any training.
While the baselines (\ghntwo, MLP and \ghnt) do no gain from ensembling, it may or may not be due to less diverse parameters. Evaluation of ensembles of predicted parameters is challenging, because some networks with predicted parameters can be of very poor quality weakening the ensemble. Therefore, we also evaluated the ensemble after fine-tuning predicted parameters which have a smaller variance of accuracy.
\rand gains the most from ensembling, perhaps because \rand leads to more diverse parameters (see Table~\ref{tab:diversity}). However, GHN-3+SGD-1ep is still much stronger than \randnospace+SGD-1ep in terms of final performance with 48.39 vs 27.45. We believe large ensembles of GHN-3+SGD-1ep networks is a promising avenue for future research.

\begin{table}[thpb]
    \centering
    \caption{Ensembling results on ImageNet (top-1 accuracy).}
    \label{tab:ensembling}
    \begin{center}
    \begin{scriptsize}
    \begin{sc}
    \setlength{\tabcolsep}{1pt}
    \begin{tabular}{lcccc}
    \toprule
     \textbf{Method} & \textbf{Resnet-50} & \textbf{ResNet-101} & \textbf{Wide-Resnet-101} & \textbf{Ensemble} \\
     \midrule
    \ghntwo & 1.08 & 1.46 & 0.70 & 1.06 \\
     MLP & 2.52 & 1.91 & 2.51 & 2.61 \\
     \ghnt & 10.59 & 5.41 & 6.17 & 8.63 \\
     \ghnthree & 19.92 & 18.86 & 18.60 & 21.07 \\
     \randnospace+SGD-1ep & 16.63 & 20.46 & 19.97 & 27.45 \\
     GHN-3+SGD-1ep & 43.33 & 42.61 & 43.11 & 48.39 \\
    \bottomrule
    \end{tabular}
    \end{sc}
    \end{scriptsize}
    \end{center}
    \vskip -0.1in
\end{table}

\subsubsection{Comparison to FixUp}

Methods such as FixUp~\cite{zhang2019fixup} provide better initialization than \rand in some cases and so they are related to our work. We compare to FixUp in Table~\ref{tab:no_bn} using ResNet-50 without batch normalization. 
Our GHN-3 based initialization significantly improves convergence compared to FixUp and all the other initializations by a large margin (Table~\ref{tab:no_bn}). In addition, our \ghnthree has a broader scope improving initialization for the networks with batch norm as well (Table~\ref{tab:init_resnet_swint}), whereas FixUp and normalization-free networks are focused on training the networks without batch norm and do not improve results with batch norm.

\begin{table}[thpb]
    \centering
    \caption{Comparison of initialization methods for ResNet-50 without batch normalization on ImageNet after 1 epoch of training.}
    \label{tab:no_bn}
    \begin{center}
    \begin{scriptsize}
    \begin{sc}
    \begin{tabular}{lcccc}
    \toprule
     \textbf{Init. method} & \textbf{Init. time} & \textbf{1 epoch} \\
     \midrule
    \rand & 0.2 & 10.8 \\
    \grad & 1700 & 19.2 \\
    FixUp & 0.2 & 18.0 \\
    \ghntwo & 1.3 & 5.9 \\
    \ghnthree & 1.2 & \textbf{31.7} \\
    \bottomrule
    \end{tabular}
    \end{sc}
    \end{scriptsize}
    \end{center}
    \vskip -0.1in
\end{table}

\subsubsection{Choice of optimizers}

Depending on the optimizer, different initialization methods may be more performant. To study that, we train Swin-T with AdamW (as in Section~\ref{sec:full_train}) and SGD with momentum and compare our \ghnthree-based initialization to \rand after 1 epoch of training (Table~\ref{tab:adam_sgd}). The results show better performance of our \ghnthree in all cases.

\begin{table}[thpb]
    \centering
    \caption{ImageNet top-1 accuracy of Swin-T after 1 epoch for different optimizers and initializations.}
    \label{tab:adam_sgd}
    \begin{center}
    \begin{scriptsize}
    \begin{sc}
    \setlength{\tabcolsep}{2pt}
    \begin{tabular}{lcccc}
    \toprule
     \textbf{Init. method} & AdamW & SGD (lr=0.05) & SGD (lr=0.025) & SGD (lr=0.01) \\
     \midrule
    \rand & 18.4 & 12.7 & 10.6 & 6.8 \\
    \ghnthree  & \textbf{27.1} & \textbf{19.4} & \textbf{19.2} & \textbf{16.1} \\
    \bottomrule
    \end{tabular}
    \end{sc}
    \end{scriptsize}
    \end{center}
    \vskip -0.1in
\end{table}

\subsubsection{Learning rate sensitivity}
An intriguing question to ask is: does initializing ImageNet models using GHN-3 makes them more robust to hyperparameters? To verify this, we computed the fraction of networks (\%) in the PyTorch split for which the best accuracy (after 1 epoch) can be achieved using SGD with the same learning rate (Table~\ref{tab:lr_sensitivity}). We found that with the GHN-3 initialization the same learning rate could achieve strong results more frequently than with \rand, so our GHN-3 initialization can make hyperparameter tuning easier.

\begin{table}[thpb]
    \centering
    \caption{The fraction of networks (\%) in the PyTorch split for which the best accuracy (after 1 epoch) can be achieved using SGD with the same learning rate.}
    \label{tab:lr_sensitivity}
    \begin{center}
    \begin{tiny}
    \begin{sc}
    \setlength{\tabcolsep}{1pt}
    \begin{tabular}{lccc}
    \toprule
     \textbf{Init. method} & \textbf{best acc$\pm$0\%} & \textbf{best acc$\pm$1\%} & \textbf{best acc$\pm$2\%} \\
     \midrule
    \rand (lr=0.1 is best on average) & 62.2 & 75.7 & 79.7 \\
    \ghnthree (lr=0.01 is best on average) & \textbf{66.2} & \textbf{78.4} & \textbf{86.5} \\
    \bottomrule
    \end{tabular}
    \end{sc}
    \end{tiny}
    \end{center}
    \vskip -0.1in
\end{table}

\begin{figure*}[t]
\begin{center}
\begin{tabular}{ccc}
     \small \textsc{ResNet-50} & \small \textsc{ResNet-152} & \small \textsc{Swin-T} \\
     {\includegraphics[width=0.3\textwidth]{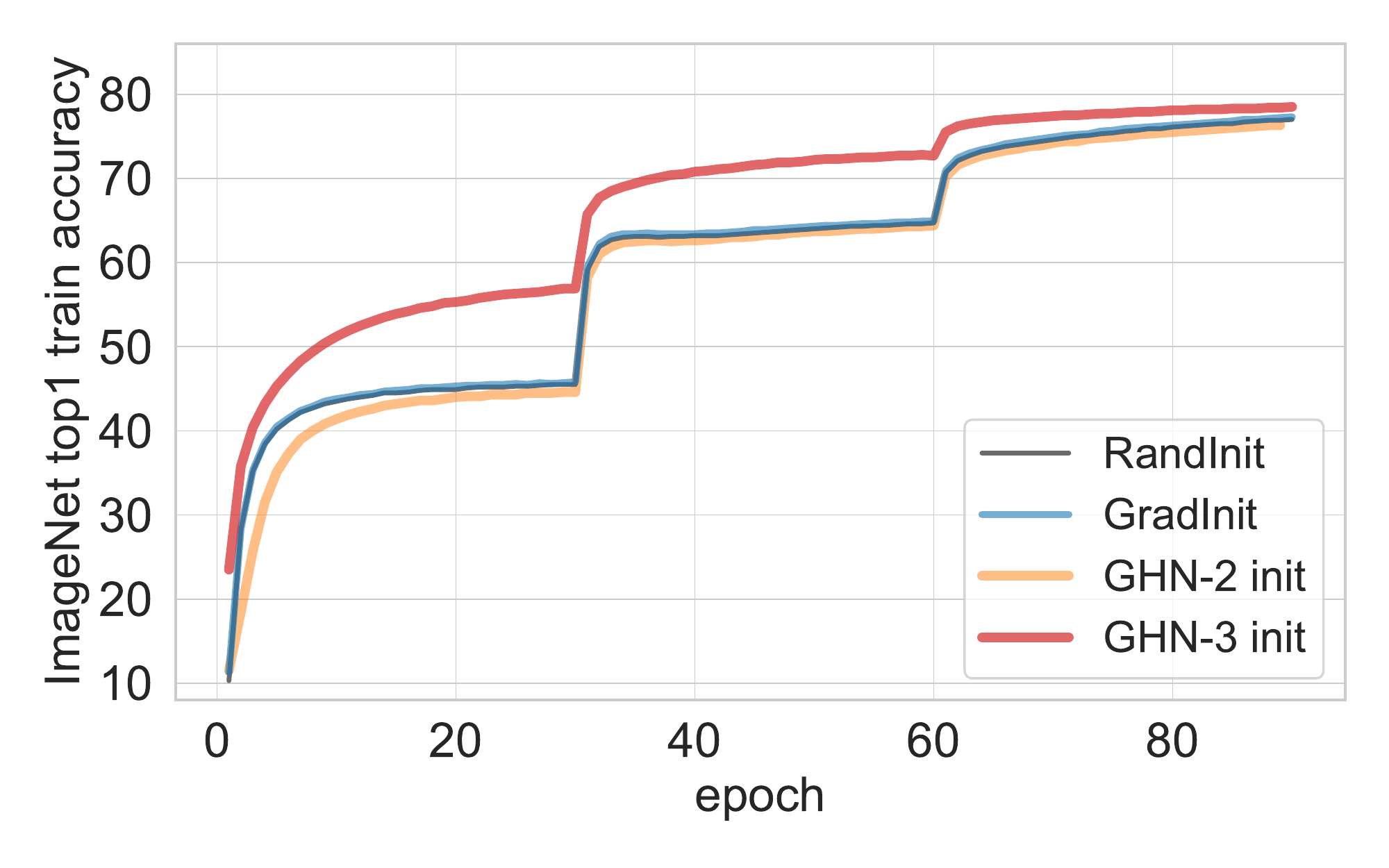}} & {\includegraphics[width=0.3\textwidth]{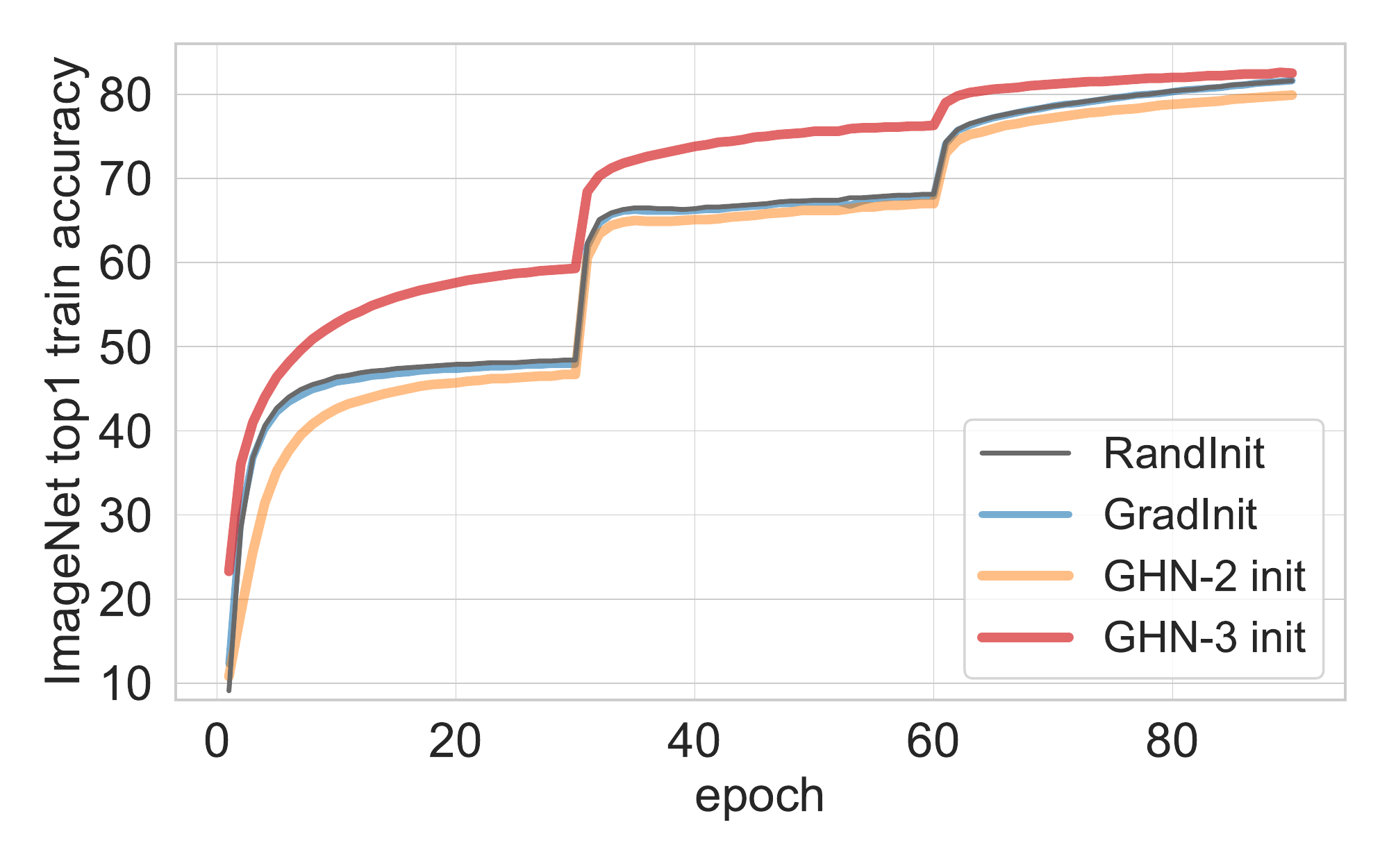}} & {\includegraphics[width=0.3\textwidth]{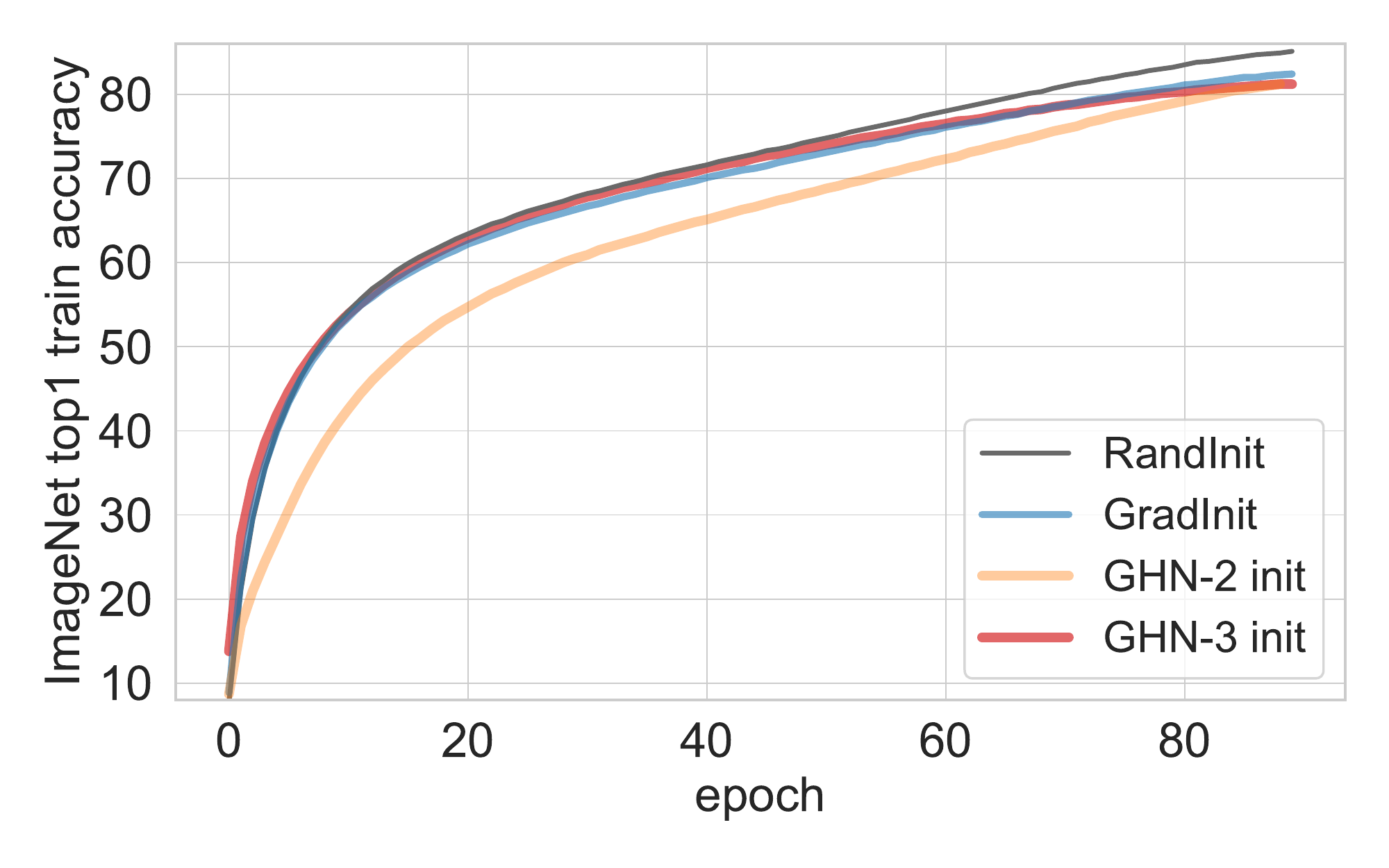}} \\
     {\includegraphics[width=0.3\textwidth]{resnet50_train_val.pdf}}
      & 
     {\includegraphics[width=0.3\textwidth]{resnet152_train_val.pdf}}
     & {\includegraphics[width=0.3\textwidth]{swint_train_val.pdf}} \\
\end{tabular}
\vskip -0.2in
\caption{Training (\textbf{top row}) and validation (\textbf{bottom row}) accuracy curves for ResNet-50, ResNet-152 and Swin-T on ImageNet. Standard learning rate schedules are used for ResNets~\cite{he2016deep} (decay every 30 epochs) and Swin-T (cosine decay)~\cite{liu2021swin}. These plots supplement Fig.~\ref{fig:val_curves} in the main text.}
\label{fig:train_val_curves}
\end{center}
\vskip -0.2in
\end{figure*}

\begin{figure*}[htpb]
\vskip 0.2in
\begin{center}
\begin{tabular}{cc}
\includegraphics[width=0.45\textwidth]{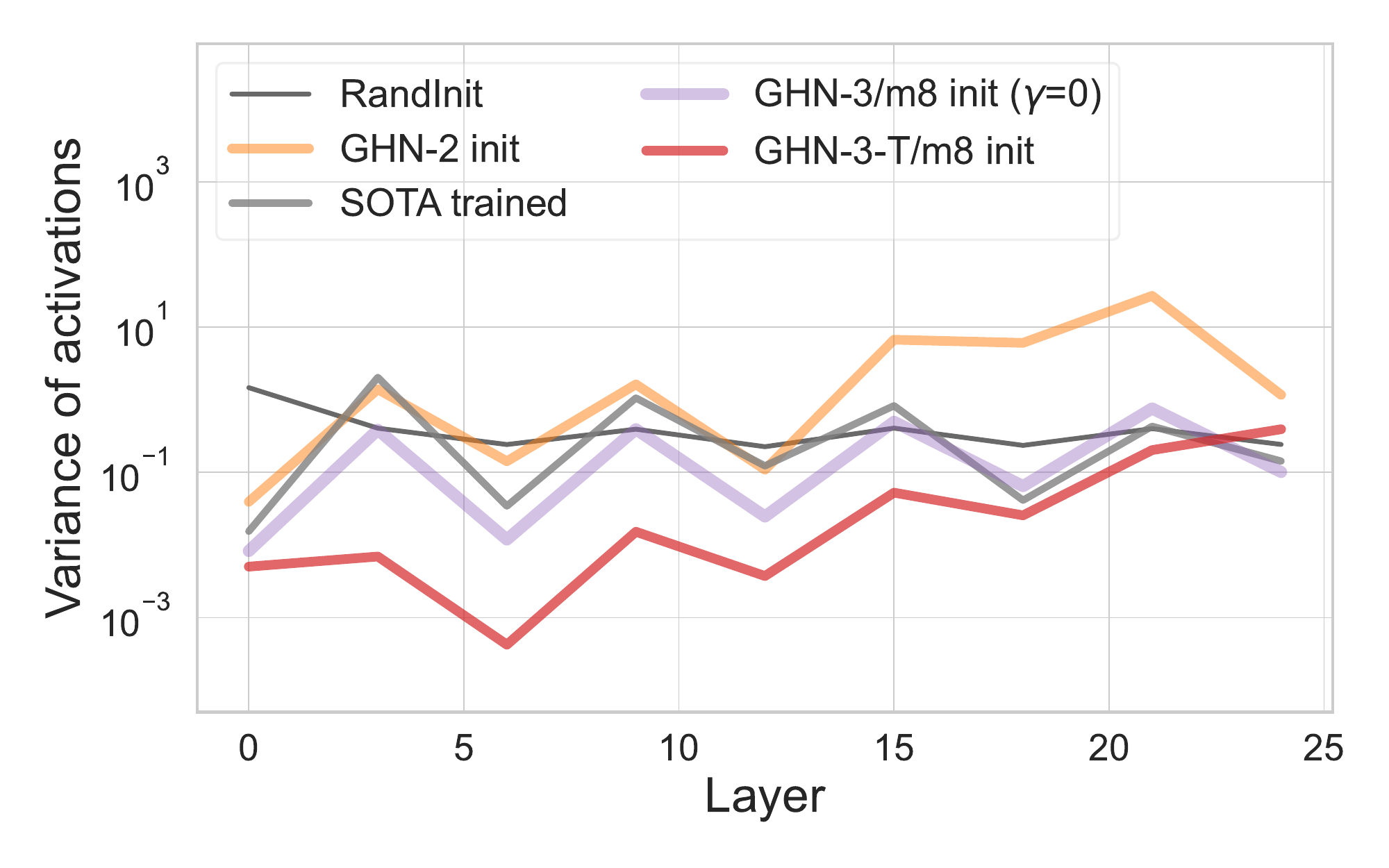} &
\includegraphics[width=0.45\textwidth]{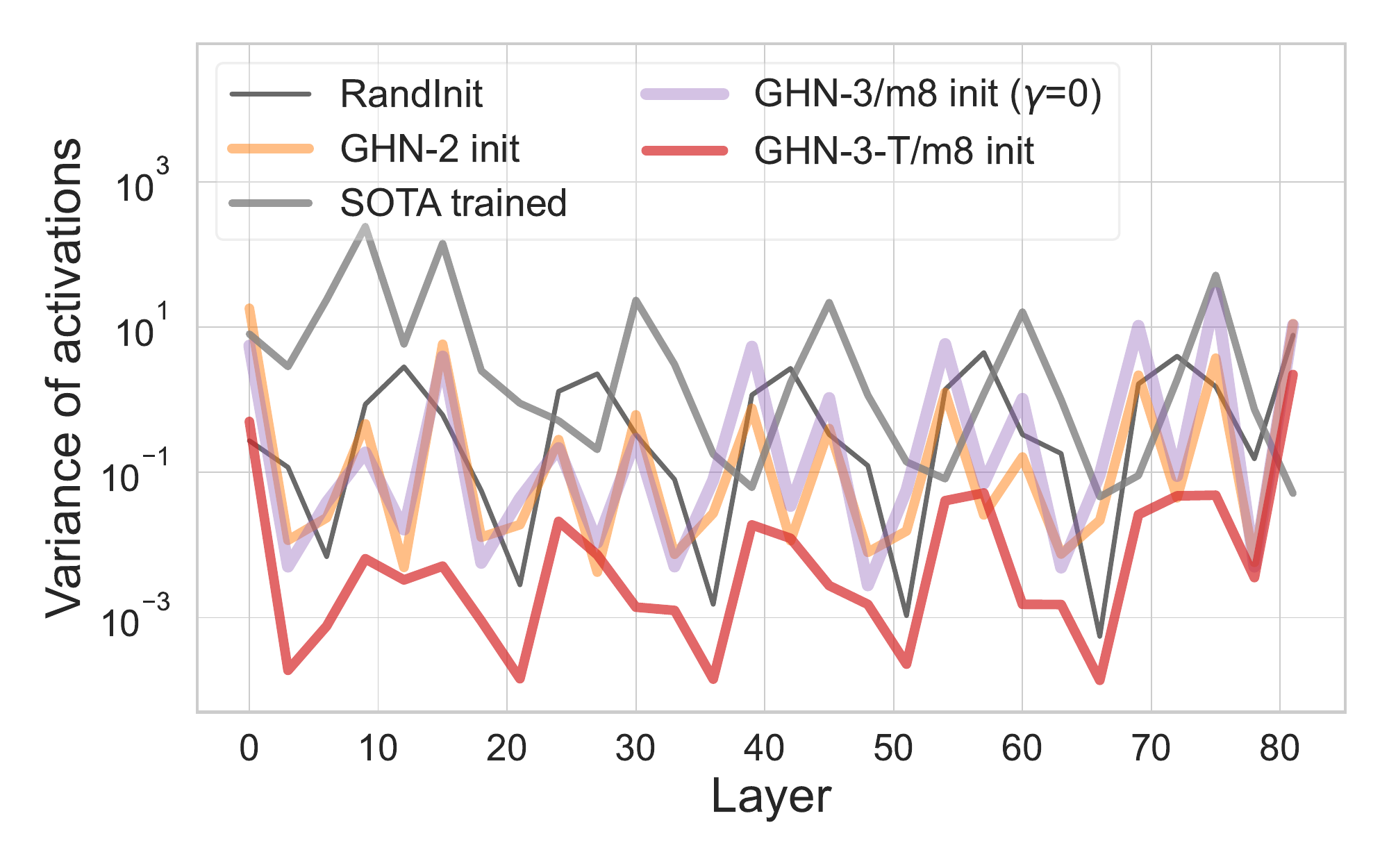}
\end{tabular}
\vskip -0.2in
\caption{Variance of activations in ViT-B/16 (\textbf{top}) and EfficientNet-B0 (\textbf{bottom}) when initialized with different methods. As in Fig.~\ref{fig:val_curves}-bottom, we compute the variance of activations only after convolutional or linear layers. To avoid clutter, we plot the variance only for every third layer. These plots supplement the discussion in Section~\ref{sec:qualitative} of the main text and are also related to the distribution shift issue discussed in Section~\ref{apdx:distr_shift}.}
\label{fig:variances_vit}
\end{center}
\vskip -0.2in
\end{figure*}

\subsection{Neural Architecture Search}
\label{apdx:nas}

\subsubsection{Related Work on Meta-learning and NAS}
\label{apdx:related-work-nas}

Several works extended hypernetworks to generalize to both architectures and datasets~\cite{lian2020towards,elsken2020meta,chen2020catch}. These works are generally based on combining meta-learning, e.g. MAML~\cite{finn2017model}, and differentiable NAS, e.g. DARTS~\cite{liu2018darts}.
These works focus on NAS, i.e. finding a strong architecture on a given dataset. To achieve this goal, they significantly constrain the flexibility of architectures for which they predict parameters.
While GHNs can also perform NAS by predicting parameters for candidate architectures, the capabilities of GHNs extend beyond NAS. 
At the same time, GHNs are not explicitly trained to perform NAS, therefore high performance of GHNs in NAS is not expected.

\subsubsection{NAS results}
While GHNs are not explicitly trained to rank architectures according to their performance of training with SGD, we verify if \ghnthree improves on \ghntwo and how it compares to strong NAS methods~\cite{abdelfattah2021zero}. We compute Kendall’s Tau rank correlation between the ImageNet accuracies obtained using predicted parameters and trained from \rand with SGD on the \dataset and \textsc{PyTorch} splits.
We first consider \randnospace+SGD-1ep as the ground truth ranking. In that case, our \ghnxl/m16 provides better rank correlation than GHN-2 on both \dataset (0.24 vs 0.16) and \textsc{PyTorch} (0.26 vs 0.08), see Table~\ref{tab:nas}.
Our GHN-3 also beats all the methods from~\cite{abdelfattah2021zero} for \dataset (0.24 vs 0.23), but is inferior on \textsc{PyTorch} (0.26 vs 0.40).
However, when we consider SOTA training as the ground truth ranking, all GHNs provide no or low negative correlation. \randnospace+SGD-1ep also provides only 0.07 correlation with SOTA training. In contrast, the methods from~\cite{abdelfattah2021zero} perform well (best correlation is 0.34) showing that GHNs may not be the best approach for NAS.
The ``jacobcov'' and ``grasp'' methods from~\cite{abdelfattah2021zero} failed for many architectures due to missing gradient implementations for some layers, so their results are not reported in some cases.

\begin{table}[t]
    \centering
    \caption{NAS results measured as Kendall's Tau Rank Correlation between ImageNet accuracies and accuracies obtained by GHNs or metrics obtained by other methods from~\cite{abdelfattah2021zero}.
    }
    \label{tab:nas}
    \vskip 0.15in
    \begin{center}
    \begin{scriptsize}
    \begin{sc}
    \setlength{\tabcolsep}{2pt}
    \begin{tabular}{lc|cc} 
        \toprule
        \textbf{Method} & \multicolumn{1}{c|}{\textbf{\dataset}} & \multicolumn{2}{c}{\textbf{PyTorch}} \\
        & \randnospace+SGD-1ep & \randnospace+SGD-1ep & SOTA \\
        \midrule
        rand ranking  & -0.00 & 0.03 & 0.07 \\
        GHN-2/m8  & 0.16 & 0.08 & -0.08 \\
        GHN-3-T/m8  & 0.22 & 0.14 & -0.05 \\
        GHN-3-XL/m16  & \textbf{0.24} & \textbf{0.26} & -0.06 \\
        \randnospace+SGD-1ep  & 1.00 & 1.00 & 0.07 \\
        
        \midrule
        \multicolumn{4}{l}{\textbf{Methods tuned for NAS from}~\cite{abdelfattah2021zero}} \\
        gradnorm  & 0.13 & 0.19 & \textbf{0.34} \\
        snip  & 0.16 & 0.20 & \textbf{0.34} \\
        fisher  & 0.12 & 0.17 & 0.24 \\
        jacobcov  & 0.23 & $-$ & $-$ \\
        plain  & 0.20 & -0.23 & -0.13 \\
        synflow-bn  & -0.31 & -0.09 & -0.08 \\
        synflow  & -0.02 & \textbf{0.42} & 0.24 \\
        
        \bottomrule
    \end{tabular}
    \end{sc}
    \end{scriptsize}
    \end{center}
    \vskip -0.1in
\end{table}

\subsection{Limitations}
\label{apdx:limitations}

We highlight the following main limitations of our work. Note that these limitations are expected, since GHNs are not specifically trained to address them.

\begin{itemize}
    \item \textbf{NAS.} As discussed in Section~\ref{apdx:nas}, GHNs are not explicitly trained to perform the NAS task and in our experiments underperformed compared to other NAS methods. At the same time, our GHN-3-XL/m16 outperformed GHN-2 and a smaller scale GHN-3 indicating the promise of large-scale GHNs for NAS.
    
    \item \textbf{Diversity of predicted parameters.} Currently, GHNs are not generative, so they deterministically predict relatively similar parameters for different architectures (Section~\ref{sec:qualitative}). This also explains low gains of ensembling in Table~\ref{tab:ensembling}. Making GHNs generative is an interesting avenue for future research.
    
    \item \textbf{Generalization.} For some architectures with unusual connectivity and unseen layers, the GHNs might not predict useful parameters. There are several generalization axes (evaluated in detail in the GHN-2 paper by~\citet{knyazev2021parameter}) and since GHNs are neural networks trained on a limited set of training architectures, some generalization gap is likely to be present. See Section~\ref{apdx:distr_shift} for the related discussion and results.
    
    \item \textbf{SOTA final performance.} To achieve SOTA ImageNet results using our initialization, the \ghnthree-based initialization generally requires about the same number of training epochs. However, compared to \rand our initialization leads to faster convergence, so that the performance slightly worse than SOTA is achieved relatively fast (see Fig.~\ref{fig:train_val_curves}, Table~\ref{tab:init_resnet_swint}). Therefore, our approach may be more optimal with low computational budgets.
    
\end{itemize}

\begin{table*}[t]
    \centering
    \caption{ImageNet top-1 accuracy for 74 PyTorch architectures using the \ghnthree-based initialization (\ghnxl/m16) vs \rand. These results are also available at \url{https://github.com/SamsungSAILMontreal/ghn3/blob/main/ghn3_results.json}. Top-10 architectures in each column are bolded, top-1 is bolded and underlined. Sorted by the last column. \textsuperscript{*}Kendall’s Tau correlation (see the last row) in each column is computed w.r.t. the accuracies in the first column. See Fig.~\ref{fig:pytorch} for histograms.
    }
    \label{tab:ghn3-full-results}
    \begin{center}
    \begin{tiny}
    \begin{sc}
    \begin{tabular}{lccccc}
    \toprule
    \textbf{Architecture} & \textbf{\ghnthree} & \textbf{GHN-3+SGD-1ep} & \textbf{GHN-3+SGD-10ep} & \textbf{\randnospace+SGD-1ep} & \textbf{SOTA} \\
    \midrule
alexnet & 0.10 & 11.87 & 36.96 & 6.24 & 56.52 \\
squeezenet1-0 & 0.11 & 4.06 & 35.12 & 5.04 & 58.09 \\
squeezenet1-1 & 0.10 & 3.33 & 33.99 & 3.73 & 58.18 \\
shufflenet-v2-x0-5 & 0.12 & 19.82 & 37.37 & 14.61 & 60.55 \\
mobilenet-v3-small & 0.11 & 15.74 & 40.31 & 17.42 & 67.67 \\
mnasnet0-5 & 0.13 & 26.90 & 47.41 & 17.51 & 67.73 \\
vgg11 & 0.12 & 22.91 & 54.03 & 12.15 & 69.02 \\
shufflenet-v2-x1-0 & 0.10 & 30.44 & 49.82 & 20.54 & 69.36 \\
resnet18 & \textbf{2.21} & \textbf{41.63} & 56.67 & 22.39 & 69.76 \\
googlenet & \textbf{2.12} & 36.64 & 57.41 & 21.07 & 69.78 \\
vgg13 & 0.12 & 24.04 & 40.47 & 12.14 & 69.93 \\
vgg11-bn & 0.11 & 24.91 & 56.28 & 12.27 & 70.37 \\
mnasnet0-75 & 0.13 & 29.80 & 52.77 & 18.21 & 71.18 \\
vgg13-bn & 0.09 & 26.99 & 57.86 & 10.89 & 71.59 \\
vgg16 & 0.12 & 21.38 & 54.31 & 9.17 & 71.59 \\
mobilenet-v2 & 0.12 & 27.87 & 51.37 & 22.32 & 71.88 \\
vgg19 & 0.09 & 16.31 & 51.83 & 6.06 & 72.38 \\
regnet-x-400mf & 0.10 & 30.14 & 54.15 & 17.07 & 72.83 \\
shufflenet-v2-x1-5 & 0.07 & 31.43 & 53.81 & 22.76 & 73.00 \\
resnet34 & \textbf{1.70} & 41.25 & 59.64 & 21.23 & 73.31 \\
vgg16-bn & 0.11 & 22.67 & 57.66 & 12.06 & 73.36 \\
mnasnet1-0 & 0.08 & 28.79 & 54.42 & 21.09 & 73.46 \\
mobilenet-v3-large & 0.14 & 18.39 & 43.15 & 23.93 & 74.04 \\
regnet-y-400mf & 0.12 & 30.88 & 56.04 & 20.30 & 74.05 \\
vgg19-bn & 0.11 & 20.42 & 56.64 & 11.96 & 74.22 \\
densenet121 & 0.14 & 31.91 & 59.26 & \textbf{25.13} & 74.43 \\
regnet-x-800mf & 0.12 & 31.41 & 58.31 & 21.18 & 75.21 \\
densenet169 & 0.15 & 31.53 & 61.92 & \textbf{24.56} & 75.60 \\
vit-b-32 & 0.13 & 6.30 & 27.18 & 8.00 & 75.91 \\
resnet50 & \textbf{20.00} & \textbf{43.69} & 59.23 & 18.19 & 76.13 \\
shufflenet-v2-x2-0 & 0.14 & 32.05 & 56.78 & 22.35 & 76.23 \\
regnet-y-800mf & 0.09 & 31.91 & 59.23 & 23.45 & 76.42 \\
mnasnet1-3 & 0.12 & 29.34 & 57.90 & 23.82 & 76.51 \\
densenet201 & 0.14 & 32.13 & 62.26 & \textbf{25.03} & 76.90 \\
vit-l-32 & 0.12 & 6.53 & 26.35 & 7.25 & 76.97 \\
regnet-x-1-6gf & 0.28 & 34.35 & 59.96 & 23.51 & 77.04 \\
densenet161 & 0.20 & 33.75 & \textbf{63.72} & \textbf{25.12} & 77.14 \\
inception-v3 & 0.37 & 40.61 & 62.57 & 8.11 & 77.29 \\
resnet101 & \textbf{18.85} & \textbf{43.15} & 60.02 & 17.45 & 77.37 \\
resnext50-32x4d & 0.16 & 32.38 & 61.89 & 20.81 & 77.62 \\
efficientnet-b0 & 0.12 & 22.83 & 53.84 & \textbf{24.64} & 77.69 \\
regnet-y-1-6gf & 0.17 & 32.35 & 62.01 & 23.36 & 77.95 \\
resnet152 & \textbf{17.07} & \textbf{43.27} & 60.02 & 16.89 & 78.31 \\
regnet-x-3-2gf & \textbf{0.87} & 35.36 & 61.89 & 22.46 & 78.36 \\
wide-resnet50-2 & \underline{\textbf{22.64}} & \underline{\textbf{44.02}} & 62.20 & 19.65 & 78.47 \\
efficientnet-b1 & 0.08 & 19.79 & 54.82 & 24.00 & 78.64 \\
wide-resnet101-2 & \textbf{18.59} & \textbf{43.11} & \textbf{62.63} & 20.77 & 78.85 \\
regnet-y-3-2gf & 0.27 & 38.02 & \textbf{64.13} & \textbf{26.72} & 78.95 \\
resnext101-32x8d & 0.16 & 35.27 & 54.15 & 21.35 & 79.31 \\
regnet-x-8gf & 0.37 & \textbf{42.01} & \textbf{64.70} & 24.54 & 79.34 \\
vit-l-16 & 0.10 & 8.92 & 27.77 & 11.15 & 79.66 \\
regnet-y-8gf & 0.14 & 41.26 & \textbf{66.01} & \underline{\textbf{28.34}} & 80.03 \\
regnet-x-16gf & \textbf{10.46} & \textbf{41.76} & \textbf{66.02} & 24.54 & 80.06 \\
regnet-y-16gf & 0.24 & \textbf{43.90} & \textbf{67.40} & \textbf{27.52} & 80.42 \\
efficientnet-b2 & 0.09 & 20.06 & 55.34 & 24.36 & 80.61 \\
regnet-x-32gf & 0.16 & 38.62 & \textbf{66.45} & \textbf{24.96} & 80.62 \\
regnet-y-32gf & 0.12 & \textbf{43.19} & \underline{\textbf{68.05}} & \textbf{27.27} & 80.88 \\
vit-b-16 & 0.12 & 8.50 & 34.91 & 11.06 & 81.07 \\
swin-t & 0.10 & 19.20 & 45.49 & 10.25 & 81.47 \\
efficientnet-b3 & 0.12 & 20.70 & 55.16 & 23.89 & 82.01 \\
convnext-tiny & 0.12 & 11.45 & 33.39 & 3.25 & 82.52 \\
swin-s & 0.09 & 17.23 & 43.58 & 9.89 & 83.20 \\
resnext101-64x4d & 0.10 & 34.29 & \textbf{63.71} & 22.36 & 83.25 \\
efficientnet-b4 & 0.12 & 17.88 & 55.10 & 22.01 & 83.38 \\
efficientnet-b5 & 0.10 & 10.00 & 46.71 & 18.44 & \textbf{83.44} \\
swin-b & 0.06 & 15.13 & 40.28 & 6.60 & \textbf{83.58} \\
convnext-small & 0.10 & 8.85 & 31.75 & 2.57 & \textbf{83.62} \\
efficientnet-b6 & 0.12 & 6.88 & 43.70 & 16.22 & \textbf{84.01} \\
convnext-base & 0.12 & 5.08 & 28.20 & 2.67 & \textbf{84.06} \\
efficientnet-b7 & 0.09 & 9.62 & 43.98 & 14.84 & \textbf{84.12} \\
efficientnet-v2-s & 0.10 & 10.58 & 47.23 & 21.14 & \textbf{84.23} \\
convnext-large & 0.10 & 5.39 & 25.66 & 2.00 & \textbf{84.41} \\
efficientnet-v2-m & 0.09 & 10.57 & 46.36 & 16.50 & \textbf{85.11} \\
efficientnet-v2-l & 0.09 & 6.43 & 43.68 & 12.60 & \underline{\textbf{85.81}} \\
\midrule
avg & 1.66$\pm$4.97 & 25.42$\pm$12.31 & 51.79$\pm$11.26 & 17.36$\pm$7.18 & 76.33$\pm$6.33 \\\
Kendall’s Tau correlation\textsuperscript{*} & 1.00 & 0.53 & 0.38 & 0.26 & -0.06 \\
 \bottomrule
    \end{tabular}
    \end{sc}
    \end{tiny}
    \end{center}
    \vskip -0.1in
\end{table*}

\end{document}

%% file: macros.tex
\newcommand{\bx}{\mathbf{x}}

\newcommand{\bz}{\mathbf{z}}

\newcommand{\std}[1]{{\tiny{$\pm$#1}}} 

\newcommand\Tstrut{\rule{0pt}{2.6ex}}         

\newcommand{\domain}{\mathcal{D}}

\newcommand{\h}{\mathbf{h}}

\newcommand{\w}{\mathbf{w}}  

\newcommand{\PLH}{{\mkern-0.1mu\times\mkern-0.1mu}}

\newcommand{\ghntwo}{\textsc{GHN-2}\xspace}
\newcommand{\ghnthree}{\textsc{GHN-3}\xspace}
\newcommand{\ghnt}{\textsc{GHN-3-T}\xspace}
\newcommand{\ghns}{\textsc{GHN-3-S}\xspace}
\newcommand{\ghnl}{\textsc{GHN-3-L}\xspace}
\newcommand{\ghnxl}{\textsc{GHN-3-XL}\xspace}
\newcommand{\rand}{\textsc{RandInit}\xspace}
\newcommand{\randnospace}{\textsc{RandInit}}
\newcommand{\grad}{\textsc{GradInit}\xspace}
\newcommand{\dataset}{\textsc{DeepNets-1M}\xspace}

\newcommand{\iidtest}{\textsc{Test}\xspace}

\newcommand{\wide}{\textsc{Wide}\xspace}
\newcommand{\deep}{\textsc{Deep}\xspace}
\newcommand{\dense}{\textsc{Dense}\xspace}
\newcommand{\bnfree}{\textsc{BN-free}\xspace}

\newcommand{\cmark}{\scriptsize\ding{51}}%
\newcommand{\xmark}{\scriptsize\ding{55}}%